\documentclass[lettersize,journal, twoside]{IEEEtran}
\usepackage[colorlinks,urlcolor=blue,linkcolor=blue,citecolor=blue]{hyperref}
\usepackage{amsmath,amsfonts}
\usepackage{algorithm}
\usepackage{textgreek}
\usepackage{array}
\usepackage[caption=false,font=normalsize,labelfont=sf,textfont=sf]{subfig}
\usepackage{textcomp}
\usepackage{stfloats}
\usepackage{url}
\usepackage{verbatim}
\usepackage{graphicx}
\usepackage{cite}
\usepackage{titlesec}
\usepackage{soul}
\titlespacing*{\subsection}{0pt}{1.25ex plus 1ex minus .2ex}{0.75ex plus .2ex}
\titlespacing*{\subsubsection}{0pt}{0.25ex plus 1ex minus .2ex}{0.25ex plus .2ex}

\usepackage{orcidlink}
\usepackage{caption}

\usepackage{multirow}
\usepackage[T1]{fontenc}
\usepackage[utf8]{inputenc}

\usepackage{algpseudocode}

\usepackage{booktabs,siunitx,array,threeparttable, makecell}

\begin{document}

\title{SDEC: Semantic Deep Embedded Clustering}

\author{\IEEEauthorblockN{Mohammad Wali Ur Rahman\,\orcidlink{0000-0002-5009-221X}\IEEEauthorrefmark{1}, 
Ric Nevarez\,\orcidlink{0009-0000-4097-9733}\IEEEauthorrefmark{2}, Lamia Tasnim Mim\,\orcidlink{0009-0005-7542-1453},\IEEEauthorrefmark{3}\,
Salim Hariri\,\orcidlink{0000-0003-3956-3401}\IEEEauthorrefmark{1} \textit{Senior Member, IEEE}}\\
\thanks{This work was supported in part by the National Science Foundation (NSF) projects under Grant 1624668, Grant 1921485, Grant 2413009 and WISPER Center. Support was also provided by the 2025 Technology and Research Initiative Fund/National Security Systems Initiative administered by the University of Arizona, Office of Research and Partnerships, funded under Proposition 301, the Arizona Sales Tax for Education Act, in 2000. Any opinions, findings, conclusions, or recommendations expressed in this paper are those of the author(s) and do not necessarily reflect the views of NSF. \\
\indent \IEEEauthorrefmark{1} Mohammad Wali Ur Rahman and Dr. Salim Hariri are with the Department of Electrical and Computer Engineering from the University of Arizona, Tucson, AZ 85721, USA. 
Email: \{mwrahman, hariri\}@arizona.edu. \\
\indent \IEEEauthorrefmark{2} Ric Nevarez is with Trustweb, New York, NY 10001, USA.
Email: ric@thetrustweb.com. \\
\indent \IEEEauthorrefmark{3} Lamia Tasnim Mim is with the Department of Computer Science from New Mexico State University, Las Cruces, NM 88003, USA. 
Email: lamia@nmsu.edu.
}
}
\markboth{Journal of IEEE Transactions on Big Data,~Vol.~XX, No.~X, Month~2025}%
{Rahman \MakeLowercase{\textit{et al.}}: SDEC: Semantic Deep Embedded Clustering}


\maketitle

\begin{abstract}
The high dimensional and semantically complex nature of textual Big data presents significant challenges for text clustering, which frequently lead to suboptimal groupings when using conventional techniques like k-means or hierarchical clustering. This work presents Semantic Deep Embedded Clustering (SDEC), an unsupervised text clustering framework that combines an improved autoencoder with transformer-based embeddings to overcome these challenges. This novel method preserves semantic relationships during data reconstruction by combining Mean Squared Error (MSE) and Cosine Similarity Loss (CSL) within an autoencoder. Furthermore, a semantic refinement stage that takes advantage of the contextual richness of transformer embeddings is used by SDEC to further improve a clustering layer with soft cluster assignments and distributional loss. The capabilities of SDEC are demonstrated by extensive testing on five benchmark datasets: \textit{AG News, Yahoo! Answers, DBPedia, Reuters 2,} and \textit{Reuters 5}. The framework not only outperformed existing methods with a clustering accuracy of 85.7\% on \textit{AG News} and set a new benchmark of 53.63\% on \textit{Yahoo! Answers}, but also showed robust performance across other diverse text corpora. These findings highlight the significant improvements in accuracy and semantic comprehension of text data provided by SDEC's advances in unsupervised text clustering. 
\end{abstract}


%
\IEEEpeerreviewmaketitle

\section{Introduction}
The introduction of transformer-based models, like BERT, has significantly advanced the field of natural language processing (NLP) in recent years \cite{devlin2018bert}. These models have established new standards in a number of NLP tasks, such as named entity recognition \cite{agrawal2022bert}, text classification \cite{sun2019howtofinetunebert, rahman2022bert, wolff2024enriched, rahman2023quantized}, and question answering \cite{rajpal2023bertologynavigator, zaib2021bert}. Text clustering is still a difficult task though, especially when working with big and complicated datasets. Since the ideal number of clusters or the ideal separation of data is frequently ambiguous, clustering is known to be an NP-hard problem, meaning there is no clear-cut solution or answer to the clustering problem \cite{kel2019fast, gonzalez1982computational}. Conventional clustering methods such as k-means and hierarchical clustering often fail to fully capture the semantic nuances of textual data. To address these issues, this paper presents SDEC (Semantic Deep Embedded Clustering), a novel text clustering framework that makes use of transformer embeddings and an autoencoder fine-tuning technique with specialized loss functions. 

Tasks like document sorting, topic detection, and information retrieval depend on text clustering, which organizes related texts into coherent clusters. The high dimensionality of text data, capturing semantic similarities, improving clustering accuracy, and guaranteeing scalability are some of the challenges this process must overcome. The curse of dimensionality causes traditional clustering techniques to frequently falter in high-dimensional spaces \cite{bellman1961adaptive}. Moreover, these conventional algorithms usually depend on Euclidean distances, which may fail to capture the semantic relationships among text data effectively  \cite{mikolov2013efficient}.

One major obstacle in textual data processing is the sparsity and noise. High-dimensional text data typically include a lot of irrelevant features, complicating the ability of clustering algorithms to detect meaningful patterns.  To mitigate this issue, feature extraction techniques such as Latent Semantic Analysis (LSA) \cite{deerwester1990indexing} and Term Frequency-Inverse Document Frequency (TF-IDF) \cite{salton1988term} have been used. But these methods frequently rely on human feature engineering and often fail to capture deeper semantic relationships. 

Word embeddings \cite{mikolov2013distributed} and, more recently, transformer embeddings \cite{devlin2018bert} have been used in recent methods to provide dense and semantically rich text representations. By capturing contextual information, these embeddings significantly enhance the performance of clustering algorithms. However, the complexity and variability of textual data make it difficult to cluster text data effectively even with these sophisticated embeddings. To address this, specialized techniques are required. 

To overcome these text clustering issues, we present the SDEC framework, which introduces several novel features. Specifically, we use BERT to produce dense, semantically rich transformer embeddings that we optimize with a novel mix of pooling algorithms and normalizing techniques to improve text processing. The core of SDEC is a novel semantic loss function that combines Cosine Similarity Loss and Mean Squared Error (MSE). This method increases the accuracy of clustering by using a dynamic weight allocation mechanism for these losses and refined data point assignments, while also maintaining semantic relationships during the reconstruction process of the autoencoder. Furthermore, a refinement stage for semantic clustering improves the quality of the clusters by modifying their assignments according to contextual similarities, making the clusters more accurate representations of the underlying data structure. When combined, these components support a strong clustering performance that successfully handles the nuanced complexity of textual data.

We demonstrate the effectiveness of SDEC through extensive experiments conducted on five benchmark datasets: \textit{AG News}, \textit{Yahoo! Answers}, \textit{DBPedia}, and two \textit{Reuters} datasets. On the \textit{AG News} dataset, SDEC achieved a clustering accuracy of 85.7\%, surpassing previous methods and setting a new benchmark. For the more challenging \textit{Yahoo! Answers} dataset, which comprises ten distinct clusters, our system attained a mean clustering accuracy of 53.63\%, improving upon the prior best of 52.6\%. SDEC also maintained strong performance on \textit{DBPedia, Reuters 2, }and\textit{ Reuters 5}, showcasing its generalizability across diverse datasets.

In addition to generalizability, we also address the scalability of our clustering methods. Section II reviews related work in the field, Section III describes the methodology of the SDEC framework, and Section IV details the experimental setup. Section V provides a detailed complexity analysis of the SDEC algorithm, assessing the computational costs during both the training and testing phases. Section VI presents the results, demonstrating the effectiveness of SDEC across various datasets. Finally, Section VII offers conclusions, highlighting the contributions and findings of our research. These sections together underscore SDEC's capability to handle large-scale datasets while achieving high clustering accuracy.

\section{Related Works}

\subsection{Advances in General Clustering Techniques}

Unsupervised learning relies heavily on clustering, and many techniques have been developed to increase the task's scalability and performance. Because of their ease of use and interpretability, traditional clustering algorithms like k-means \cite{lloyd1982least} and hierarchical clustering \cite{murtagh2012algorithms} have been used extensively. Density-based clustering algorithms, such as DBSCAN, HDBSCAN, DenLAC \cite{ester1996density, campello2013density, ruadulescu2021denlac} identify clusters based on local density variations and can detect arbitrarily shaped clusters and effectively managing noise. Nevertheless, these techniques frequently have trouble handling high-dimensional data and might miss intricate patterns within the data.

Researchers have proposed various advanced clustering techniques to address these limitations. For example, before clustering, spectral clustering \cite{ng2002spectral} reduces the dimensionality by using the eigenvalues of a similarity matrix. The PV-DM \cite{radulescu2020density} model learns a fixed vector for each document as a step of dimensionality reduction and combines it with word vectors to predict the next word \cite{radu2020clustering}. 

Another promising direction is community detection which characterizes communities of similar items by representing data as network rather than purely feature-space distances \cite{fortunato2016community}. Graph-based community detection methods, including modularity maximization algorithms like Louvain \cite{blondel2008fast}, partition documents into communities and model them as nodes in a similarity network. The effectiveness of community detection has been further enhanced by approaches that integrate hybrid communities-clustering graph structures \cite{ruadulescu2020enhancing}. FN-BERT-TFIDF \cite{apostol2024contcommrtd} model detects geolocation-content-based communities on Twitter using topic modeling and deep learning for real-time hazard analysis and misinformation detection.

The capacity of deep learning-based clustering techniques to extract complex representations from data has drawn much interest in recent years. Using autoencoder networks and clustering, Deep Embedded Clustering (DEC) \cite{xie2016unsupervised} learns feature representations and iteratively refines cluster assignments. Dizaji et al. \cite{ghasedi2017deep} proposed Deep Embedded Regularized Clustering (DEPICT), which employed a convolutional autoencoder (CAE) and stacked a softmax layer on the embedded features to introduce a novel clustering loss. Li et al. \cite{li2018discriminatively} presented Discriminatively Boosted Image Clustering (DBC), adopting a self-paced learning strategy where simpler instances are tackled first, then progressively introducing more complex samples. Both approaches build upon the principles of DEC but incorporate additional mechanisms, such as softmax-based clustering in DEPICT and self-paced learning in DBC, to further improve clustering performance. A reconstruction loss is added to DEC in the Improved Deep Embedded Clustering (IDEC) \cite{guo2017improved} in order to maintain the local structure of the data while clustering.

For clustering tasks, variational autoencoders (VAEs) \cite{kingma2013auto} have also been investigated. In order to enhance clustering performance, VAEs have been coupled with clustering techniques to learn probabilistic latent variable models. One such method that allows for better handling of complex data distributions is the Gaussian Mixture Variational Autoencoder (GMVAE) \cite{dilokthanakul2016deep}. It models the latent space as a mixture of Gaussians. 

Furthermore, clustering has been done using Generative Adversarial Networks (GANs) \cite{goodfellow2014generative}. By applying cluster-level constraints to the latent space, ClusterGAN \cite{mukherjee2019clustergan} combines clustering with GANs, producing better clustering outcomes.

In order to overcome the limitations of DEC, Lee et al. \cite{lee2022deep} developed a deep embedded clustering framework for mixed data that handles both numerical and categorical data while enhancing convergence stability through the use of soft-target updates. When applied to mixed data, their method outperforms conventional approaches. 

\subsection{Recent Innovations in Text Clustering Approaches}

Topic modeling has long been a foundational approach in text analysis, aiming to summarize a larger collection of documents. Classical topic modeling methods such as Latent Dirichlet Allocation (LDA) \cite{blei2003latent} and Non-negative Matrix Factorization (NMF) \cite{lee2000algorithms} have historically played a significant role in text clustering and topic detection. LDA models documents as probabilistic mixtures over latent topics, providing interpretable topic distributions. NMF offers an alternative by decomposing document-term matrices into two non-negative matrices often leading to more coherent topics. Clustering techniques have become integral to various social media analysis tasks, enabling insights into user behavior, event dynamics, and public opinion. For example, deep learning-based clustering architectures have been employed to predict audience interest in emerging news topics on social platforms. These models, either independently or combined with additional techniques such as transformer-based embeddings, have been widely applied to analyze sentiment, reputation, influence, and other social dynamics within topic-specific contexts \cite{mitroi2020sentiment, petrescu2019sentiment, petrescu2301edsa, truicua2021deep}. Studies have explored ways to enhance topic modeling outcomes by comparing different weighting schemes (e.g., TF, TF-IDF) and incorporating contextual cues and document level embedding to create better coherent topics \cite{truica2016comparing, truica2017topic, truicua2021topic}. However, they rely on bag-of-words assumptions, neglecting deep semantic and contextual information.

The high dimensionality and sparsity of text data make text clustering particularly difficult. Various feature extraction techniques have been used to adapt traditional methods, like k-means and hierarchical clustering, for text data. For example, Latent Semantic Analysis (LSA) \cite{deerwester1990indexing} and Term Frequency-Inverse Document Frequency (TF-IDF) \cite{salton1988term} have been used extensively to translate text into numerical features.

By offering dense and semantically meaningful representations of words, word embeddings like Word2Vec \cite{mikolov2013distributed}, GloVe \cite{pennington2014glove}, and FastText \cite{bojanowski2017enriching} have completely changed text clustering. Similar texts can be grouped together more easily thanks to these embeddings, which capture contextual information. Nevertheless, polysemy and capturing sentence context remain difficult tasks for word embeddings.

By offering contextual embeddings that take the entire sentence into account, transformer-based models like BERT \cite{devlin2018bert} have advanced the field even further. Text clustering is one of the many NLP tasks where BERT embeddings have proven effective. Subakti et al. \cite{subakti2022bert} investigated the use of BERT for text clustering, comparing it to more established techniques such as TF-IDF and analyzing its performance as a data representation for text clustering.

Clustering textual data is still challenging despite these developments. According to Subakti et al. \cite{subakti2022bert}, although BERT performs better than TF-IDF in the majority of metrics, the effectiveness is heavily reliant on the feature extraction and normalization techniques chosen. Utilizing a range of pooling strategies, including max pooling and mean pooling, and then varying normalization methods, they demonstrated how the clustering algorithm employed can have a substantial impact on how effective BERT representations are.

In order to address joint optimization, their study also made use of DEC and IDEC, which combine deep learning models with clustering layers. However, they only employed Mean Squared Error (MSE) losses for reconstruction, which are less useful for text data. Their detailed study concentrated mainly on comparing BERT with TF-IDF. We get around this drawback by using semantic loss functions, which are more appropriate for identifying semantic similarities in text.

To improve clustering performance on complex text datasets, our proposed framework makes use of transformer embeddings and an autoencoder fine-tuning approach with specialized semantic loss functions. In order to guarantee stable and efficient training, we also included convergence checks during the fine-tuning procedure by continuously monitoring delta values and KL divergence losses. Our approach yields improved semantic representation and clustering accuracy by combining semantic and distributional loss functions, as our benchmark dataset experiments show. By optimizing representation learning and clustering simultaneously, this integrated approach produces more cohesive and semantically significant clusters.

\begin{figure*}[!t]
    \centering
    \captionsetup{justification=centering}
    \includegraphics[width=0.80\textwidth]{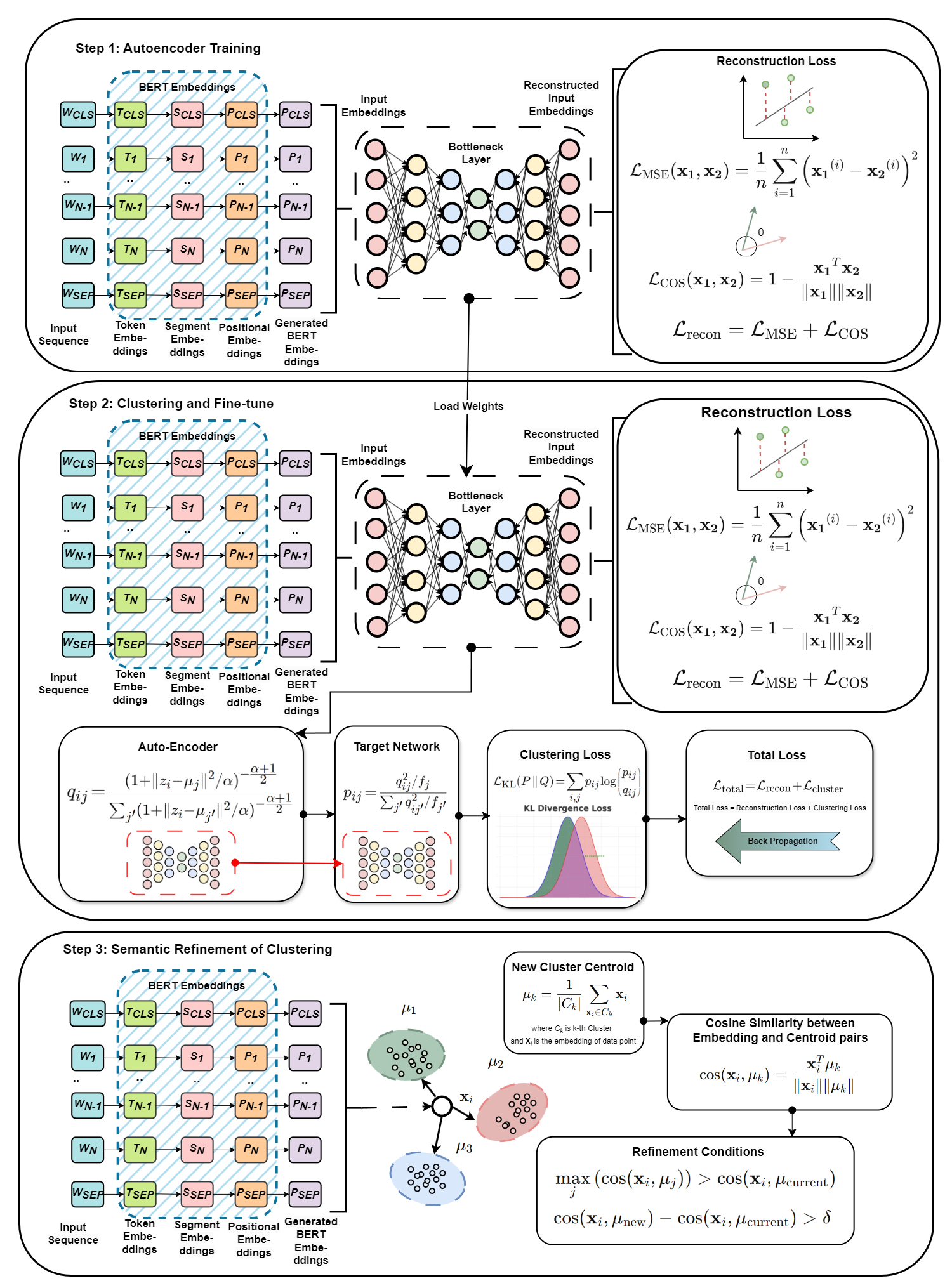}
    \caption{Semantic Deep Embedded Clustering Framework for Text Data}
    \label{fig:framework}
\end{figure*}

\section{Methodology}

Our proposed text clustering framework's general architecture makes use of transformer embeddings' strength and autoencoders' ability to fine-tune using specialized semantic loss functions. The goal of this architecture is to generate well-clustered, semantically meaningful representations of text data. Below is a summary of the architecture's main elements and procedures.

Our Semantic Deep Embedded Clustering (SDEC) framework is composed of three main stages: autoencoder-based latent space transformation, clustering with fine-tuning, and semantic refinement. To learn a compressed latent representation, text data is first transformed into BERT embeddings, which are subsequently run through an autoencoder. Through the reduction of noise and the capture of underlying semantic structures, this latent space facilitates more efficient clustering. A combination of distributional and semantic losses is used to further refine the clustering process, ensuring significant topic separations and high intra-cluster similarity. In order to produce clearly defined and semantically coherent clusters, a semantic refinement phase iteratively improves cluster assignments by adjusting cluster centroids based on contextual similarity.

Our framework for semantically enhanced SDEC text clustering with combined semantic and distributional losses fine-tuning is shown in Figure \ref{fig:framework}.

The preprocessing steps, including text cleaning, lemmatization, and tokenization, as well as the generation of BERT embeddings using different pooling and normalization strategies, are detailed in Appendix A and Appendix B, respectively. These steps ensure that the textual data is transformed into high-quality embeddings suitable for clustering.

\subsection{Autoencoder Training}

The autoencoder is an essential part of our text clustering framework. In order to reconstruct the original embeddings from this latent representation, high-dimensional BERT embeddings are compressed into a lower-dimensional latent space. This procedure makes sure that the most important characteristics of the data are kept while minimizing noise, which facilitates efficient clustering. 

An encoder and a decoder network comprise the autoencoder configuration. By compressing the input embeddings into a lower-dimensional bottleneck layer, the encoder is able to extract the most important data features. From this compressed representation, the decoder then reconstructs the original embeddings. Thus, it is imperative that the bottleneck layer maintains the information that the embeddings need for precise reconstruction and subsequent clustering.

In order to prevent overfitting, the autoencoder architecture used in this study consists of multiple layers with L2 regularization and SeLU (Scaled Exponential Linear Unit) activation. The selection of SeLU was driven by its self-normalizing characteristic, which aids in preserving stable mean and variance of neuron outputs without significant dependence on supplementary normalization layers. As a result, training remains less susceptible to both vanishing and exploding gradients \cite{klambauer2017self}. This stability, coupled with the controlled output range inherent in SeLU, minimizes large updates in the network parameters, reducing the tendency to memorize noisy patterns. When combined with L2 regularization, which directly penalizes overly large weights. SeLU thus encourages the network to focus on preserving the most salient characteristics of the input embeddings rather than learning dataset-specific patterns, ultimately yielding more robust latent representations for clustering. 

The input data is compressed by the encoder using several layers, and the original input is restored by the decoder by mirroring this process. The encoder network reduces the dimensionality of the input embeddings through successive layers with SeLU activation and L2 regularization. The final layer of the encoder is the bottleneck layer, which captures the most salient features of the input data.

The bottleneck layer serves as the most compressed representation of the input data. It is defined as follows:
\begin{equation}
\mathbf{h} = f(\mathbf{W}_h \mathbf{x} + \mathbf{b}_h),
\end{equation}
where \( \mathbf{x} \) represents the input embeddings, \( \mathbf{W}_h \) and \( \mathbf{b}_h \) are the weights and biases of the bottleneck layer, and \( f \) is the activation function.

Once the bottleneck representation is obtained, the decoder network reconstructs the original embeddings. The output layer of the decoder has a linear activation function to match the dimensionality of the input embeddings.

To improve reconstruction accuracy, a combined loss function is employed that integrates both Mean Squared Error (MSE) and Cosine Similarity Loss. This ensures that the reconstruction preserves both the magnitude (via MSE) and the semantic similarity (via cosine similarity) of the original embeddings. The Mean Squared Error (MSE) Loss measures the average squared difference between the original embeddings \( \mathbf{x}_i \) and the reconstructed embeddings \( \hat{\mathbf{x}}_i \), defined as:
\begin{equation}
\mathcal{L}_{\text{MSE}} = \frac{1}{n} \sum_{i=1}^{n} (\mathbf{x}_i - \hat{\mathbf{x}}_i)^2
\end{equation}

Additionally, the Cosine Similarity Loss measures the cosine of the angle between the original and reconstructed embeddings, ensuring that they retain their semantic similarity. It is given by:
\begin{equation}
\cos(\mathbf{x}_i, \hat{\mathbf{x}}_i) = \frac{\mathbf{x}_i \cdot \hat{\mathbf{x}}_i}{\|\mathbf{x}_i\| \|\hat{\mathbf{x}}_i\|}
\end{equation}
where \( \mathbf{x}_i \cdot \hat{\mathbf{x}}_i \) denotes the dot product, and \( \|\mathbf{x}_i\| \) and \( \|\hat{\mathbf{x}}_i\| \) denote the magnitudes of the vectors. The Cosine Similarity Loss is then given by:
\begin{equation}
\mathcal{L}_{\text{cosine}} = 1 - \cos(\mathbf{x}_i, \hat{\mathbf{x}}_i)
\end{equation}

The twofold nature of textual embedding preservation is addressed by combining MSE and Cosine Similarity Loss. In particular, MSE enforces accuracy in each embedding dimension, ensuring the reconstructed vectors retain their numerical scale and overall magnitude.  In contrast, Cosine Similarity Loss preserves the angular alignment between the original and reconstructed vectors, an essential characteristic for embeddings in natural language tasks, where direction often encodes semantic relationships (e.g., topic or sentiment). Relying on MSE alone could preserve magnitude but allow semantic drifts; conversely, using only cosine similarity could yield consistent directions but overlook scale-related nuances that can be meaningful in certain embedding dimensions.

The synergy between these two goals is further enhanced by the dynamic weighting mechanism. The MSE loss is given a correspondingly higher weight at each training step if the autoencoder has a harder time matching the scale of the embeddings (higher MSE). On the other hand, the model allocates more capacity to directional vector alignment if maintaining the semantic orientation (i.e., cosine similarity) turns into the bottleneck. The network won't overcommit to one loss component while neglecting the other thanks to this adaptive emphasis. Rather, it responds to the current reconstruction difficulty by continuously balancing the two goals, which speeds up convergence and improves final embeddings.

To optimize the model effectively, the weights between these two losses are dynamically adjusted based on their relative magnitudes, ensuring that the model focuses on whichever loss term is more dominant during training:
\begin{equation}
\mathcal{L}_{\text{recon}} = \textit{weight}_{\text{MSE}} \times \mathcal{L}_{\text{MSE}} + \textit{weight}_{\text{cosine}} \times \mathcal{L}_{\text{cosine}}
\end{equation}
where the weights are computed as:
\begin{equation}
\textit{weight}_{\text{MSE}} = \frac{\mathcal{L}_{\text{MSE}}}{\mathcal{L}_{\text{MSE}} + \mathcal{L}_{\text{cosine}} + \epsilon}
\end{equation}
and
\begin{equation}
\textit{weight}_{\text{cosine}} = \frac{\mathcal{L}_{\text{cosine}}}{\mathcal{L}_{\text{MSE}} + \mathcal{L}_{\text{cosine}} + \epsilon}
\end{equation}
where \( \epsilon \) is a small constant to prevent division by zero. This combined semantic loss function ensures that the reconstructed embeddings maintain both semantic alignment (through cosine similarity) and their original magnitude and scale (through MSE). The dynamic weight allocation allows the model to adaptively focus on the more challenging aspect of reconstruction during training.

The autoencoder is fitted to the training set of data during the training phase, and its performance is verified on a separate test set. The goal is to make sure the autoencoder learns a concise and useful representation of the input data, as this representation can be used for clustering. 

The ability of an autoencoder to learn a lower-dimensional representation of high-dimensional data makes it important for clustering tasks. This representation reduces noise and makes clustering easier by capturing the data's key characteristics. More specifically, for accurate clustering and reconstruction, the bottleneck layer provides a compact representation that retains the most important data. 

By employing the autoencoder, we ensure the effective compression and reconstruction of the BERT embeddings, providing a solid foundation for the subsequent clustering phases. As such, the autoencoder plays a crucial role in enhancing the accuracy and performance of our text clustering framework. 

\subsection{Clustering Layer}

Our framework's clustering layer transforms the input sample (feature) into a soft label that indicates the likelihood that the sample will fall into each cluster. The similarity between the embedded point and the cluster center determines this probability. By integrating the clustering process within the neural network, both clustering and representation learning are optimized simultaneously, ensuring more cohesive and semantically meaningful clusters. Figure \ref{fig:cluster_layer} shows the SDEC network structure with the cluster layer. The clustering layer outputs a probability distribution (soft labels) across all clusters for each input sample. Hard cluster labels \mbox{(\(y_i\))} are subsequently assigned by taking the \(\arg\max\) of these probabilities.

\begin{figure}[t]
    \centering
    \captionsetup{justification=centering}
    \label{fig:cluster_layer}
    \includegraphics[width=0.98\columnwidth]{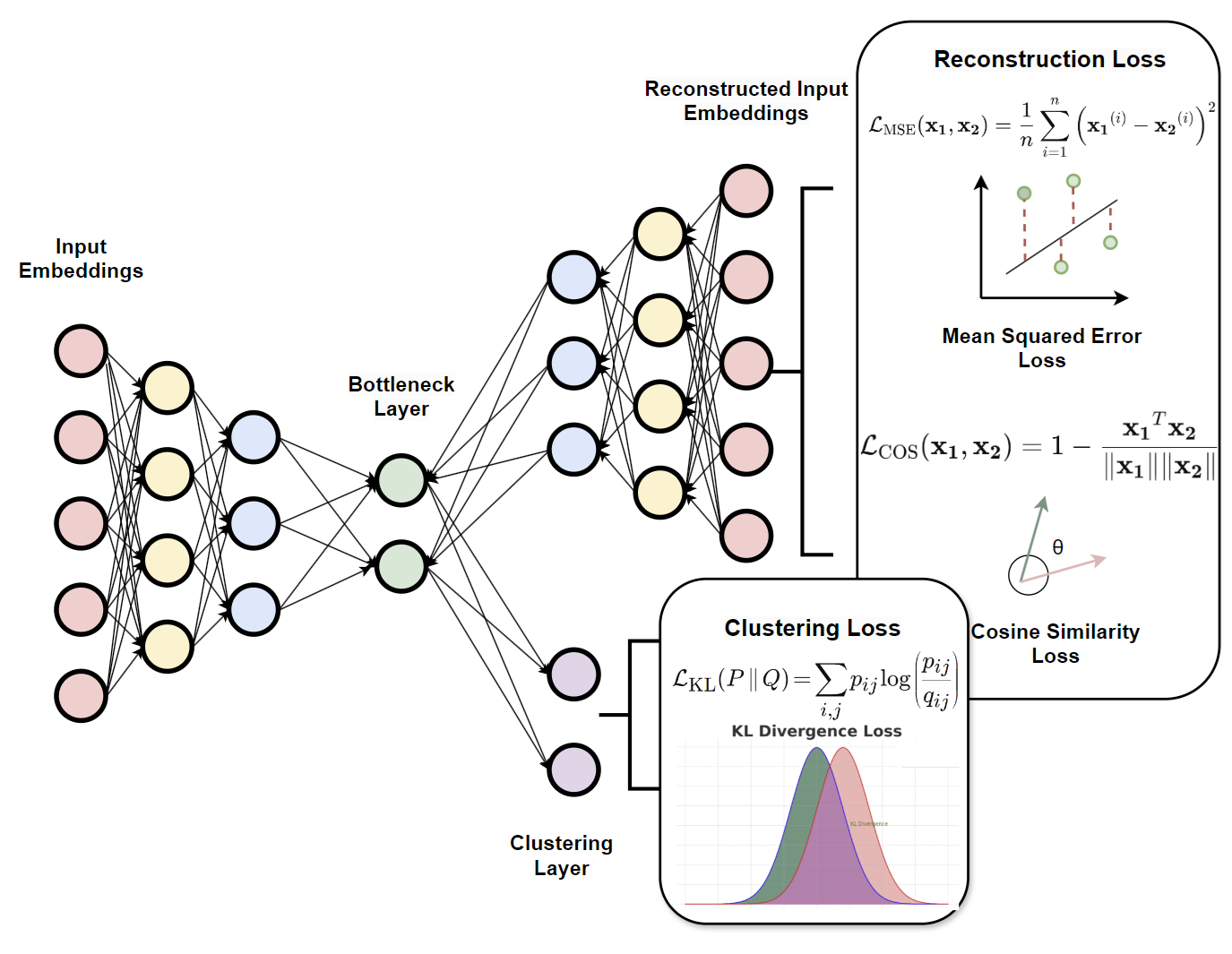}
    \caption{SDEC Network Structure with Cluster Layer}
    \label{fig:cluster_layer}
\end{figure}

A customized Keras layer is used to implement this clustering mechanism. This clustering layer acts as a soft-assignment layer based on the Student's t-distribution. During training, it computes the soft assignments for every sample, initializes the cluster centers, and updates them as the model learns. The clustering layer initializes the cluster centers \(\mathbf{\mu}_j\) using the K-means++ procedure, ensuring that the initial centers are well-distributed in the latent space \cite{lloyd1982least}. 

For each input sample \(\mathbf{z}_i\) in the latent space, the soft assignment \(q_{ij}\) to cluster \(j\) is calculated using the Student's t-distribution as a kernel:
\begin{equation}
q_{ij} = \frac{(1 + \frac{\|\mathbf{z}_i - \mathbf{\mu}_j\|^2}{\alpha})^{-\frac{\alpha+1}{2}}}{\sum_{j'} (1 + \frac{\|\mathbf{z}_i - \mathbf{\mu}_{j'}\|^2}{\alpha})^{-\frac{\alpha+1}{2}}},
\end{equation}
where \(\alpha\) is the degrees of freedom of the Student's t-distribution, \(\mathbf{z}_i\) is the latent representation of sample \(i\), and \(\mathbf{\mu}_j\) is the cluster center \(j\) \cite{maaten2008visualizing}. The target distribution \(p_{ij}\) is then computed to enhance cluster purity by emphasizing high-confidence assignments:
\begin{equation}
p_{ij} = \frac{q_{ij}^2 / \sum_i q_{ij}}{\sum_{j'} (q_{ij'}^2 / \sum_i q_{ij'})},
\end{equation}
where \(q_{ij}\) is the soft assignment from the previous step \cite{lee2022deep}.

The clustering layer employs the Kullback-Leibler (KL) divergence loss to measure the difference between the target distribution \(p\) and the predicted distribution \(q\):
\begin{equation}
\mathcal{L}_{\text{KL}} = \sum_{i} \sum_{j} p_{ij} \log \frac{p_{ij}}{q_{ij}},
\end{equation}
where \(p_{ij}\) is the target distribution and \(q_{ij}\) is the predicted distribution \cite{kingma2013auto}. 

Final hard cluster assignments for each data sample are determined by selecting the cluster index with the highest soft assignment probability:

\begin{equation}
y_i = \arg\max_j q_{ij}
\end{equation}

For optimization, the model uses Stochastic Gradient Descent (SGD) with a learning rate of 0.01 and momentum of 0.9 \cite{kingma2014adam}. The total loss for the entire network is defined as
\begin{equation}
\mathcal{L}_{\text{total}} = \mathcal{L}_{\text{cluster}} + \mathcal{L}_{\text{recon}},
\end{equation}
where \(\mathcal{L}_{\text{recon}}\) is the combined semantic loss from the reconstruction process, and \(\mathcal{L}_{\text{cluster}}\) corresponds to the KL divergence loss from the clustering layer.

Soft target updates are performed by iteratively updating the target distribution \(p\) based on the current soft assignments \(q\). This strategy refines the clustering results and improves cluster purity. During training, an iterative process updates both the model parameters via the combined loss function and the cluster centers using the predicted soft assignments. Convergence is tracked using the KL divergence and delta label changes, ensuring stability and efficiency in the clustering process.

Consistent with standard deep clustering literature, the number of clusters \( k \) is treated as a priori knowledge and corresponds directly to the known number of categories in each dataset. If the true number of clusters is unknown, it can be estimated using methods such as the elbow method, silhouette analysis, or eigengap heuristics \cite{rousseeuw1987silhouettes}.

Our method incorporates the clustering layer into the same neural network used for representation learning, ultimately enabling the reconstruction objective and the clustering objective to reinforce one another. By carefully combining these processes within a single framework, we obtain more cohesive and semantically meaningful clusters, as the network can capture both the essential data features and their underlying group structures.

In practice, simultaneously training the autoencoder and the clustering layer can lead to competing gradients, where the reconstruction objective seeks to retain as much information as possible while the clustering objective forces separation of latent representations. Training them together often causes numeric instability or forces premature convergence to suboptimal cluster centroids. By first allowing the autoencoder to learn a stable low-dimensional representation, we ensure that subsequent clustering is performed on embeddings that accurately capture semantic relationships. By using a phased approach, conflicting optimization signals are avoided and more consistent and reliable clustering results are obtained.

\subsection{Semantic Refinement of Clustering}

To improve the coherence of cluster assignments based on semantic similarity, the SDEC (Semantic Deep Embedded Clustering) algorithm has a refinement phase. In the refinement step, data points are reassigned to clusters if their similarity to a new cluster centroid is greater than their current clusters. This technique guarantees that semantically related points are grouped together and enhances cluster purity.

Let \( \mathbf{X} = \{\mathbf{x}_1, \mathbf{x}_2, \dots, \mathbf{x}_n \} \) be the set of BERT embeddings, where \( \mathbf{x}_i \in \mathbb{R}^d \) is the \( d \)-dimensional embedding of the \( i \)-th data point, and \( C = \{ c_1, c_2, \dots, c_k \} \) be the set of \( k \) cluster centroids, with initial labels assigned as \( \mathbf{y} = \{y_1, y_2, \dots, y_n\} \). The refinement process consists of the following steps:
\titlespacing*{\subsubsection}{0pt}{0.75ex plus 1ex minus .2ex}{0.25ex plus .2ex}
\subsubsection{Computation of Cluster Centroids}

For each cluster \( c_j \), we compute the cluster centroid \( \mathbf{\mu}_j \in \mathbb{R}^d \), which is the mean of all embeddings that belong to cluster \( c_j \). Mathematically, the centroid \( \mathbf{\mu}_j \) is given by:
\begin{equation}
\mathbf{\mu}_j = \frac{1}{|C_j|} \sum_{\mathbf{x}_i \in C_j} \mathbf{x}_i
\end{equation}
where \( |C_j| \) is the number of points in cluster \( c_j \), and \( \mathbf{x}_i \in C_j \) represents the data points assigned to cluster \( c_j \).
\titlespacing*{\subsubsection}{0pt}{0.75ex plus 1ex minus .2ex}{0.25ex plus .2ex}
\subsubsection{Semantic Similarity and Reassignment}
For each point \( \mathbf{x}_i \), its similarity to its current cluster centroid \( \mathbf{\mu}_{y_i} \) is computed using cosine similarity:
\begin{equation}
\text{Sim}(\mathbf{x}_i, \mathbf{\mu}_{y_i}) = \frac{\mathbf{x}_i \cdot \mathbf{\mu}_{y_i}}{\|\mathbf{x}_i\| \|\mathbf{\mu}_{y_i}\|}
\end{equation}

We also compute the cosine similarities of \( \mathbf{x}_i \) with all other cluster centroids \( \mathbf{\mu}_j \), for \( j = 1, \dots, k \). The point is reassigned to a new cluster \( c_{j^\ast} \) if the similarity to the new centroid \( \mathbf{\mu}_{j^\ast} \) exceeds the similarity to the current centroid by a predefined threshold \( \lambda \):
\begin{equation}
j^\ast = \arg\max_j \, \text{Sim}(\mathbf{x}_i, \mathbf{\mu}_j)
\end{equation}

The point \( \mathbf{x}_i \) is reassigned to \( c_{j^\ast} \) if:
\begin{equation}
\text{Sim}(\mathbf{x}_i, \mathbf{\mu}_{j^\ast}) - \text{Sim}(\mathbf{x}_i, \mathbf{\mu}_{y_i}) > \lambda
\end{equation}

The algorithm refines the initial cluster labels \( y_i \) to new labels \( y_i^{\text{new}} \) based on the computed similarities and the threshold \( \lambda \). The refined labels \( y_i^{\text{new}} \) are used to improve cluster consistency by grouping semantically similar points together.

The threshold \( \lambda \) acts as a sensitivity parameter that determines how large the similarity difference needs to be for a point to be reassigned. A higher threshold results in fewer reassignments, whereas a lower threshold leads to more aggressive refinement.

When working with high-dimensional embeddings, like BERT embeddings, this refinement technique is especially beneficial because the semantic structure of the data may not be adequately represented by the initial clustering methods. We can make sure that points are assigned to the most semantically appropriate cluster by refining clusters based on semantic similarity. 

This robust method works well for high-dimensional feature spaces because cosine similarity is used to guarantee that the refinement is independent of the size of the embeddings and only considers their directional alignment. 

\section{Experiments}

\subsection{Exploratory Data Analysis}

This subsection outlines the datasets utilized in our clustering experiments. The choice of datasets was guided by the need for diversity in topic coverage and data scale, ranging from thousands to over a million samples, to ensure robust testing of the SDEC framework.
\titlespacing*{\subsubsection}{0pt}{0.25ex plus 1ex minus .2ex}{0.25ex plus .2ex}
\subsubsection{AG News Dataset}
The \textit{AG News} dataset \cite{zhang2015character} consists of news articles divided into four major categories: World, Sports, Business, and Sci/Tech. We used 5,600 articles for training and 120,000 for testing, maintaining balance across categories to ensure fairness in clustering evaluations.

\subsubsection{Yahoo! Answers Dataset}
The \textit{Yahoo! Answers} dataset \cite{yahooanswers}, with its diverse array of user-generated questions and answers, spans ten different categories. This dataset provides a substantial challenge due to its size and variety, with 60,000 pairs for training and 1,400,000 for testing.

\subsubsection{DBPedia Dataset}
Derived from Wikipedia, the \textit{DBPedia} dataset \cite{auer2007dbpedia} categorizes structured data into fourteen clusters, ranging from geographical entities to cultural works. We employed 70,000 samples for training and 560,000 for testing.

\subsubsection{Reuters Dataset}
We utilized two subsets of the \textit{Reuters-21578} corpus, focusing on major business and economic categories. The \textit{Reuters 2} subset includes categories \textit{"earn"} and \textit{"acq"}, with 1,243 training and 4,972 testing articles. The more complex \textit{Reuters 5} subset expands to include five major categories including \textit{"earn", "acq", "money-fx", "grain"} and \textit{"crude"}, with 1,442 training and 6,487 testing articles. Notably, the \textit{Reuters 5} dataset is heavily imbalanced, with the \textit{"earn"} and \textit{"acq"} categories comprising the vast majority of articles \cite{reuters21578}.

These datasets, detailed in Table \ref{tab:description_datasets}, were preprocessed to remove non-textual elements and ensure clean text for clustering. The variety and scale of these datasets are critical for assessing the generalizability and effectiveness of clustering algorithms.

\begin{table}[h!]
\centering
\caption{Description of Datasets}
\label{tab:description_datasets}
\resizebox{\columnwidth}{!}{%
\begin{tabular}{|c|c|c|c|c|c|}
\hline
\textbf{Dataset} & \makecell{\textbf{Train} \\ \textbf{Samples}} & \makecell{\textbf{Test} \\ \textbf{Samples}} & \textbf{Clusters}  & \makecell{\textbf{Max/Min} \\ \textbf{Cluster} \\ \textbf{Size Ratio}} & \makecell{\textbf{Clustering} \\ \textbf{Complexity}}\\ \hline
AG News        & 5,600     & 120,000     & 4 & 1 : 1 & Moderate \\
\hline
\makecell{{Yahoo!} \\ {Answers}} & 60,000    & 1,400,000   & 10 & 1 : 1 & High \\
\hline
DBPedia        & 70,000    & 560,000     & 14 & 1 : 1 & High \\
\hline
Reuters 2      & 1,243     & 4,972       & 2 & 1.7 : 1 & Low\\
\hline
Reuters 5      & 1,442     & 6,487       & 5 & 13.4 : 1 & High \\
\hline
\end{tabular}
}
\end{table}

\begin{table*}[h]
\centering
\caption{Execution Time and Scalability Metrics for SDEC}
\label{tab:execution_time}
\begin{tabular}{|l|c|c|c|c|c|c|c|c|}
\hline
\textbf{Dataset} & \makecell{\textbf{Training} \\ \textbf{Samples}} & \makecell{\textbf{Training} \\ \textbf{Time (s)}} & \makecell{\textbf{Testing} \\ \textbf{Samples}} & \makecell{\textbf{Testing} \\ \textbf{Time (s)}} & \makecell{\textbf{No. of} \\ \textbf{Clusters}} & \makecell{\textbf{Avg. No. of} \\ \textbf{BERT Tokens} \\ \textbf{Combined}} & \makecell{\textbf{Avg. No. of} \\ \textbf{BERT Tokens} \\ \textbf{Training}} & \makecell{\textbf{Avg. No. of} \\ \textbf{BERT Tokens} \\ \textbf{Testing}} \\ \hline
AG News        & 5000  & 1647.22 & 120000  & 809.55  & 4  & 44.8   & 44.81 & 44.59  \\ \hline
Yahoo! Answers & 60000 & 2132.05 & 1400000 & 1407.7  & 10 & 145.64 & 146.02 & 144.89 \\ \hline
Reuters 2      & 1243  & 1151.113 & 4972    & 571.19  & 2  & 111.56 & 111.77 & 109.68 \\ \hline
Reuters 5      & 1442  & 1339.4  & 6487    & 849.58  & 5  & 126.51 & 126.73 & 124.56 \\ \hline
DBPedia        & 70000 & 3121.6  & 560000  & 1487.83 & 14 & 72.37  & 72.23  & 72.64  \\ \hline
\end{tabular}
\end{table*}

\subsection{Experiment Settings}

To provide clarity on our experimental methodology, all hyperparameter settings and training configurations, including details on autoencoder architecture, clustering fine-tuning, and cluster refinement strategies, are provided in Appendix C. These settings define the training conditions under which our SDEC clustering framework was evaluated.

\subsection{Evaluation Metrics}

In this subsection, we discuss the evaluation metrics used to assess the performance of our clustering model. The primary metrics considered are Clustering Accuracy (ACC), Normalized Mutual Information (NMI), and Adjusted Rand Index (ARI). Each of these metrics provides a different perspective on the quality of the clustering results, and their mathematical definitions are provided below.

\subsubsection{Clustering Accuracy (ACC)}

Clustering Accuracy (ACC) measures the percentage of correctly assigned labels after the best mapping between the predicted cluster labels and the true labels is found. It is defined as follows:

\begin{equation}
\text{ACC} = \max_{m} \frac{\sum_{i=1}^{n} \mathbf{1}\{y_i = m(c_i)\}}{n}
\end{equation}

where \( y_i \) is the true label, \( c_i \) is the cluster label, \( m \) ranges over all possible one-to-one mappings between clusters and labels, \( \mathbf{1}\{\cdot\} \) is the indicator function, and \( n \) is the total number of samples. The accuracy measures how well the clustering model can reproduce the true labels after the optimal mapping.

\subsubsection{Normalized Mutual Information (NMI)}

Normalized Mutual Information (NMI) measures the amount of shared information between the clustering results and the true labels. It normalizes the mutual information score to scale between 0 and 1. The NMI is defined as:

\begin{equation}
\text{NMI}(Y, C) = \frac{2 \cdot I(Y; C)}{H(Y) + H(C)}
\end{equation}

where \( I(Y; C) \) is the mutual information between the true labels \( Y \) and the cluster assignments \( C \), and \( H(Y) \) and \( H(C) \) are the entropies of \( Y \) and \( C \) respectively. NMI evaluates how much information about the true labels is captured by the cluster assignments, with 1 indicating perfect correlation and 0 indicating no correlation.

\subsubsection{Adjusted Rand Index (ARI)}

The Adjusted Rand Index (ARI) measures the similarity between the predicted and true cluster assignments, adjusting for chance grouping. The ARI is defined as:

\begin{equation}
\text{ARI} = \frac{\text{RI} - \mathbb{E}[\text{RI}]}{\max(\text{RI}) - \mathbb{E}[\text{RI}]}
\end{equation}

where \( \text{RI} \) is the Rand Index, which calculates the number of pairs of samples that are either correctly assigned to the same cluster or correctly assigned to different clusters, and \( \mathbb{E}[\text{RI}] \) is the expected value of the Rand Index for random cluster assignments. ARI ranges from -1 to 1, where 1 indicates perfect agreement, 0 indicates random labeling, and negative values indicate worse-than-random labeling.

By taking into account both the agreement with true labels and the accuracy of cluster assignments, these metrics collectively offer a thorough assessment of the clustering performance.

\begin{figure*}[!t]
    \centering
    \subfloat[Training Time vs. Training Samples\label{fig:sdec_training}]{%
        \includegraphics[width=0.45\linewidth]{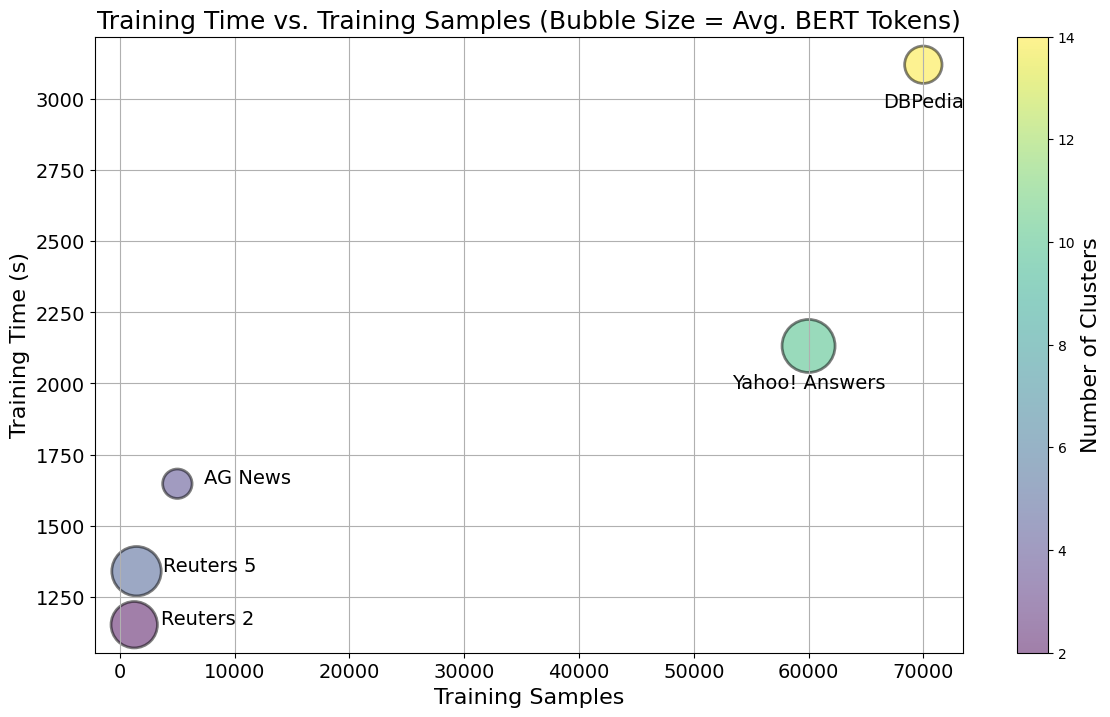}
    }
    \hfill
    \subfloat[Testing Time vs. Testing Samples\label{fig:sdec_testing}]{%
        \includegraphics[width=0.45\linewidth]{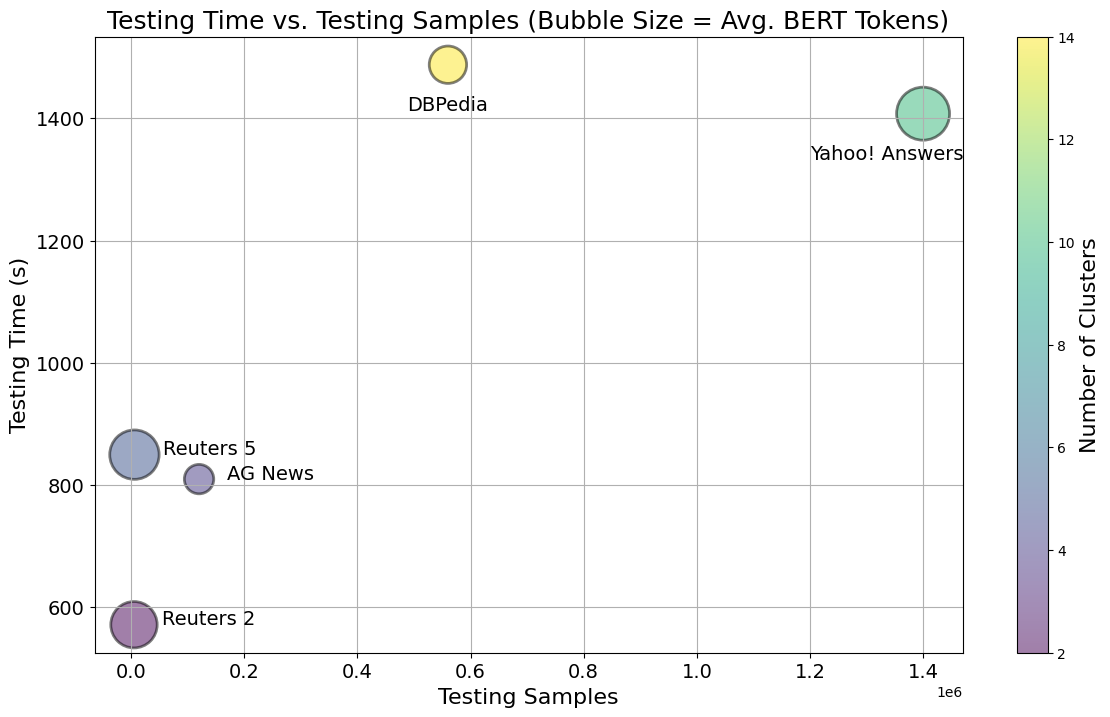}
    }
    \caption{Scalability of the SDEC Algorithm. Bubble size corresponds to the average number of BERT tokens, and color indicates the number of clusters.}
    \label{fig:sdec_scalability}
\end{figure*}

\section{Computational Complexity}

In this section, we analyze the computational complexity of the proposed SDEC framework, which consists of three primary components: the autoencoder with a combined loss (MSE and cosine similarity), the clustering process and the semantic refinement. The complexity is presented for both the training and testing phases.

Preprocessing and converting text to BERT embeddings are preparatory steps essential for the subsequent stages of the framework. While these steps have computational costs, they are generally executed once per dataset and do not scale with the core parameters of the clustering algorithm. Therefore, their complexity is not detailed here but is acknowledged to be part of the preprocessing overhead.
\subsection{Training Complexity}

The training phase of SDEC involves the autoencoder's forward and backward passes, combined loss computation, clustering, and semantic refinement. The overall complexity is derived by considering the following components:

\subsubsection{Autoencoder with Combined Loss}
The forward and backward propagation through the autoencoder, which consists of $L$ layers, has a complexity of:
\[
O\left( 2 \times \sum_{i=1}^{L} n_{i-1} \times n_i \right)
\]
where $n_i$ represents the number of neurons in the $i$-th layer. 

For the combined loss:
\begin{itemize}
    \item \textbf{Mean Squared Error (MSE) loss}: The complexity of MSE loss is $O(B \times d)$, where $B$ is the batch size and $d$ is the dimensionality of the embeddings.
    \item \textbf{Cosine Similarity loss}: The complexity of the cosine similarity loss is also $O(B \times d)$.
\end{itemize}

Hence, the total complexity of the autoencoder with the combined loss function becomes:
\[
O\left( 2 \times \sum_{i=1}^{L} n_{i-1} \times n_i + B \times d \right)
\]

\subsubsection{Clustering and Semantic Refinement}
The clustering phase in SDEC includes:
\begin{itemize}
    \item \textbf{K-means initialization}: This has a complexity of $O(n_b \times k \times d)$, where $n_b$ is the number of data points, $k$ is the number of clusters, and $d$ is the dimensionality of the embeddings 
    \item \textbf{KL Divergence loss}: Calculating the KL divergence loss between the soft cluster assignments has a complexity of $O(i \times n_b \times k \times d)$ where $i$ represents the number of iterations for clustering convergence.
    \item \textbf{Cosine Similarity for cluster refinement}: The semantic refinement process based on cosine similarity also contributes $O(n_b \times k \times d)$.
\end{itemize}

Thus, the total complexity for clustering and refinement becomes:
\[
O(n_b \times k \times d + i \times n_b \times k \times d)
\]

\subsubsection{Total Training Complexity}
The total computational complexity for the training phase, combining both the autoencoder and the clustering process, is:
\[
O\left( E \times \left( \sum_{i=1}^{L} n_{i-1} \times n_i + B \times d + n_b \times k \times d \times \left(i+1\right)\right) \right)
\]
where $E$ is the number of epochs, $B$ is the batch size, $n_b$ is the number of data points, $k$ is the number of clusters, and $i$ is the number of clustering iterations.

\subsection{Testing Complexity}
In the testing phase, we do not compute backpropagation or the loss functions. The testing complexity consists of the following components:

\subsubsection{Autoencoder Forward Pass}
The forward pass through the autoencoder has a complexity of:
\[
O\left( \sum_{i=1}^{L} n_{i-1} \times n_i \right)
\]

\subsubsection{Clustering Assignments}
During the testing phase, only the clustering assignments are computed. The complexity of assigning points to clusters based on cosine similarity with cluster centroids is:
\[
O(n_b \times k \times d)
\]

\subsubsection{Total Testing Complexity}
Thus, the total computational complexity during testing is:
\[
O\left( \sum_{i=1}^{L} n_{i-1} \times n_i + n_b \times k \times d \right)
\]

The total complexity of SDEC during training is dominated by the autoencoder training, the combined loss calculation, and the clustering with semantic refinement. During testing, the complexity is significantly reduced as only the forward pass and clustering assignments are computed. The overall complexity is scalable and efficient for large-scale text clustering tasks.

\subsection{Execution Time and Scalability of SDEC}

\begin{figure*}[!t]
    \centering
    \captionsetup{justification=centering}
    \includegraphics[width=0.98\textwidth]{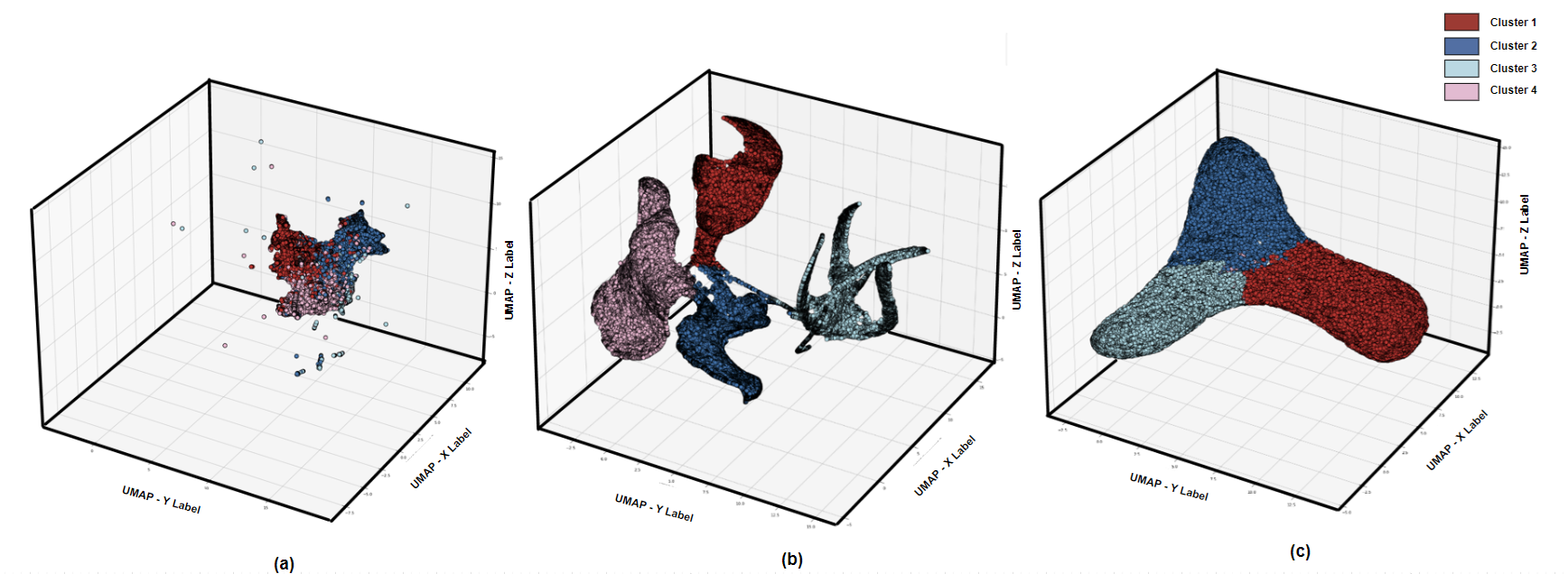}
    \caption{Visual Illustration of Step-by-Step Improvements in Clustering Performance}
    \label{fig:cluster_results}
\end{figure*}

This section presents an analysis of the SDEC (Semantic Deep Embedded Clustering) algorithm's scalability and execution time on various datasets. The number of samples, training time, testing time, and average number of BERT tokens processed during training and testing for different datasets are summarized in Table \ref{tab:execution_time}. Additionally, bubble charts in Figure \ref{fig:sdec_scalability} show the average number of BERT tokens and clusters for each dataset, as well as the relationship between training/testing time and sample sizes. 

The scalability of the SDEC algorithm in relation to the number of samples and execution time is demonstrated by the bubble charts. The average number of BERT tokens per dataset is represented by the size of the bubbles, and the number of clusters is indicated by their color.

As shown in the table and visualized in the left chart of Figure \ref{fig:sdec_scalability}, the training time increases with the number of training samples across all datasets. \textit{DBPedia} and \textit{Yahoo! Answers}, which have the highest number of training samples (70,000 and 60,000, respectively), also require the longest training times. This is consistent with the linear dependence of training complexity on the number of samples, as derived in the complexity analysis. The bubble sizes indicate that \textit{DBPedia} and \textit{Yahoo! Answers} also process the most BERT tokens, which further contributes to the increased training time.

Similar trends are observed for testing time in the right chart of Figure \ref{fig:sdec_scalability}. \textit{DBPedia} and \textit{Yahoo! Answers}, with the largest testing samples (560,000 and 1,400,000, respectively), have the highest testing times. Smaller datasets like \textit{Reuters 2} and \textit{Reuters 5} exhibit significantly shorter testing times, which is expected given their much smaller number of samples. The average number of BERT tokens per dataset also influences testing time, as seen by the varying bubble sizes.

Execution time is also dependent on the number of clusters. Larger datasets with many clusters, 14 for \textit{DBPedia} and 10 for \textit{Yahoo! Answers}, take longer to process than smaller datasets with only two clusters, like \textit{Reuters 2}. The scalability analysis demonstrates the flexibility of the SDEC algorithm in practical applications by indicating that it can handle a broad range of datasets with different sample sizes, cluster counts, and BERT token lengths.

\section{Results}

\subsection{Step-by-Step Clustering Improvements}

In this subsection, we present outcomes from three distinct experimental setups on a randomly selected run of the \textit{AG News} dataset. To emphasize the improvements made through various stages of our methodology, the results are presented through evaluation metrics and visualizations.

Figure \ref{fig:cluster_results} shows the clustering results in three stages: (a) initial clustering using an autoencoder with K-Means and MSE as the reconstruction loss, (b) clustering with a DEC layer using MSE for reconstruction and KL divergence for clustering loss without fine-tuning, and (c) our proposed SDEC clustering system with enhanced semantic loss (Cosine Similarity Loss + MSE), fine-tuning of the entire network for improved cluster assignments, and further refinement of clustering based on semantic contextual information.

We note that there is little discernible separation between the clusters in the first stage (Figure \ref{fig:cluster_results} (a)) where the data samples are mainly mixed together. The assessment metrics, which show the adjusted Rand index (ARI) at 43.2, normalized mutual information (NMI) at 50.5, and accuracy (ACC) at 56.34, are indicative of this \cite{strehl2002cluster}. The autoencoder's use of MSE as the reconstruction loss and the clustering algorithm K-Means fail to adequately capture the underlying structure of the data.

When the DEC layer is added, we observe some improvements in the second stage (Figure \ref{fig:cluster_results} (b)). The graph's middle region is one area where the clustering assignments are still unclear, despite the fact that the clusters are now more separated than they were previously. This stage produces an accuracy (ACC) of 69.71, an NMI of 53.5, and an ARI of 50.2.  Although there is a notable improvement, the clustering results still leave room for further enhancement.

\begin{figure*}[!t]
    \centering
    \subfloat[AG News Performance (Sorted by Accuracy)\label{fig:ag_performance}]{%
        \includegraphics[width=0.45\textwidth]{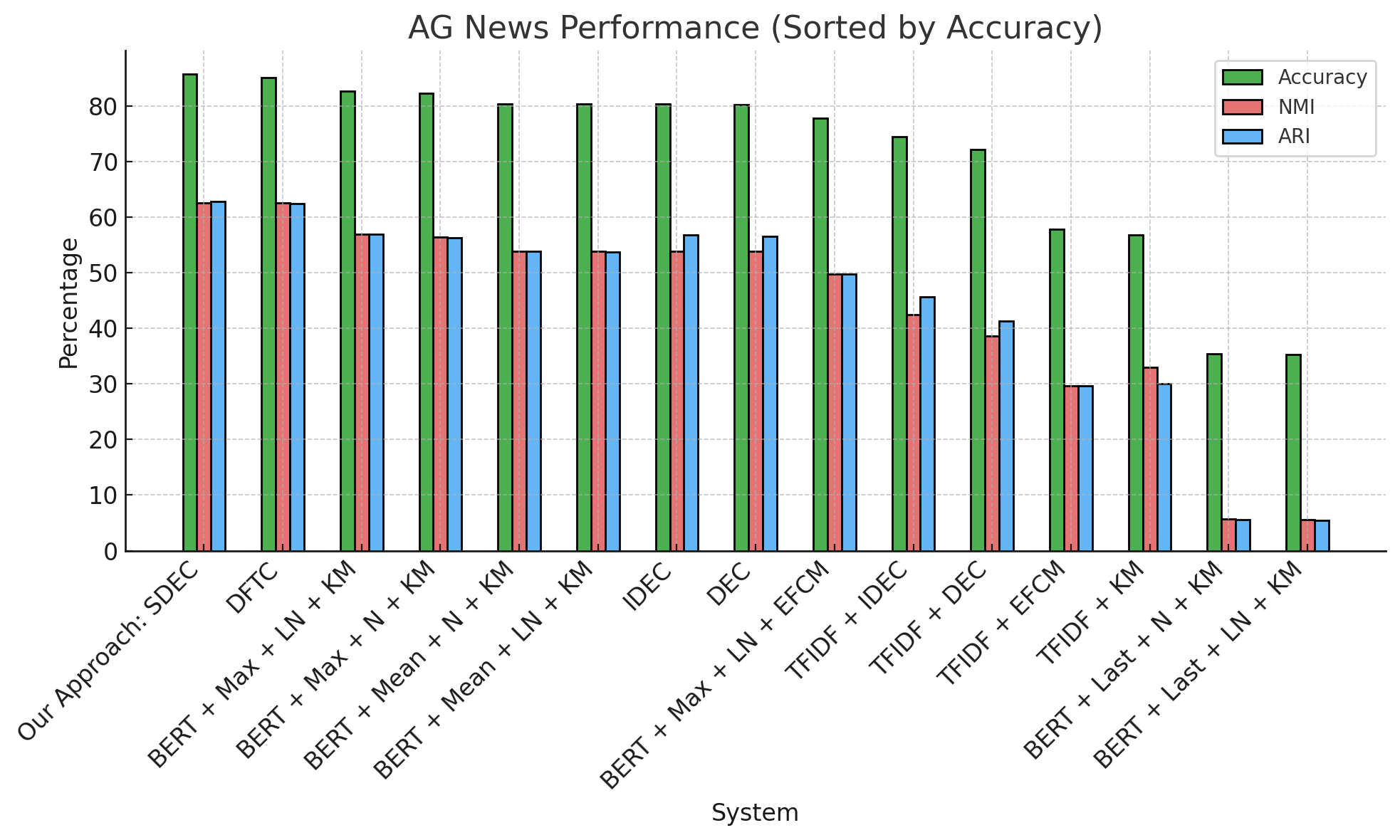}
    }
    \hfill
    \subfloat[Yahoo! Answers Performance (Sorted by Accuracy)\label{fig:yahoo_performance}]{%
        \includegraphics[width=0.45\textwidth]{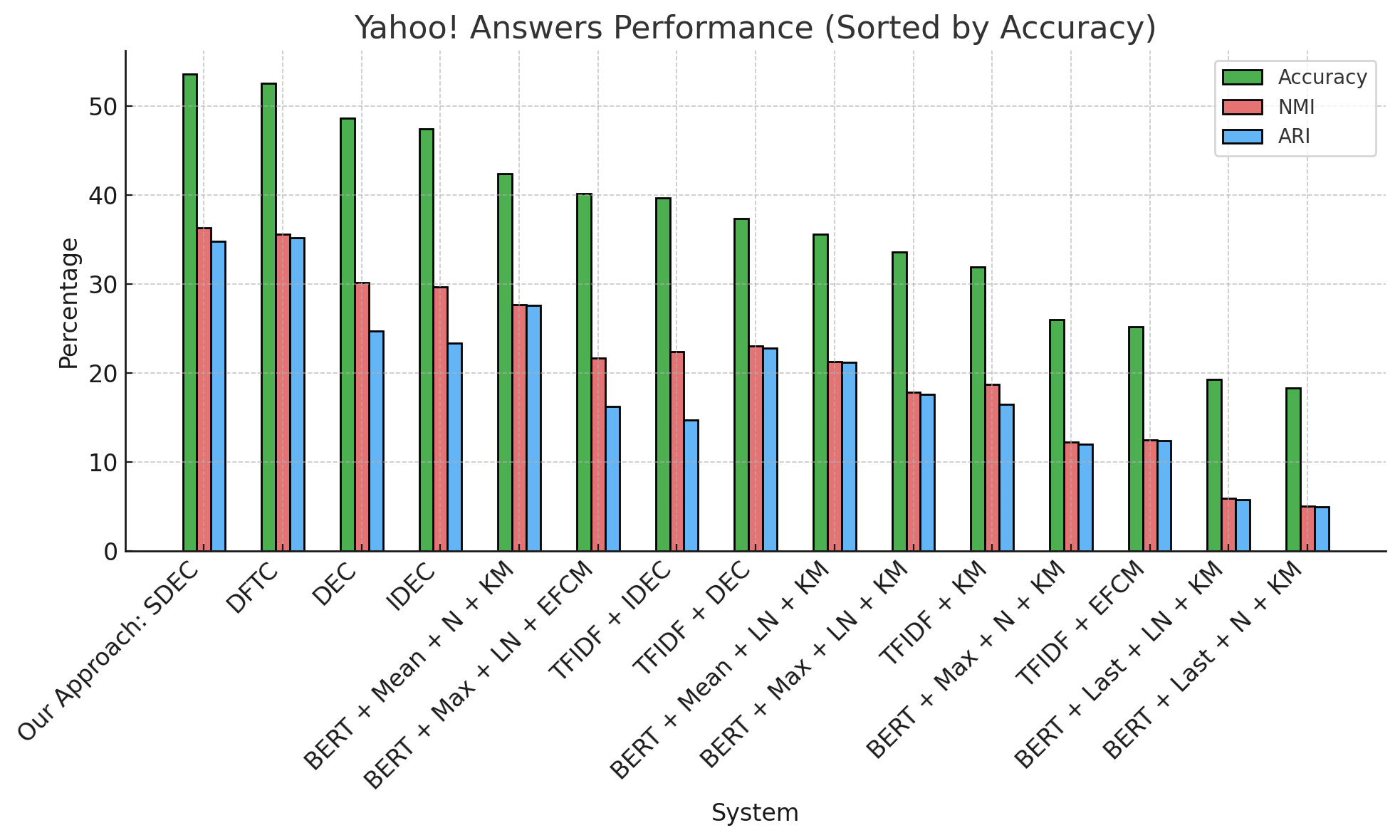}
    }
    \\
    \subfloat[DBPedia Performance (Sorted by Accuracy)\label{fig:dbpedia_performance}]{%
        \includegraphics[width=0.45\textwidth]{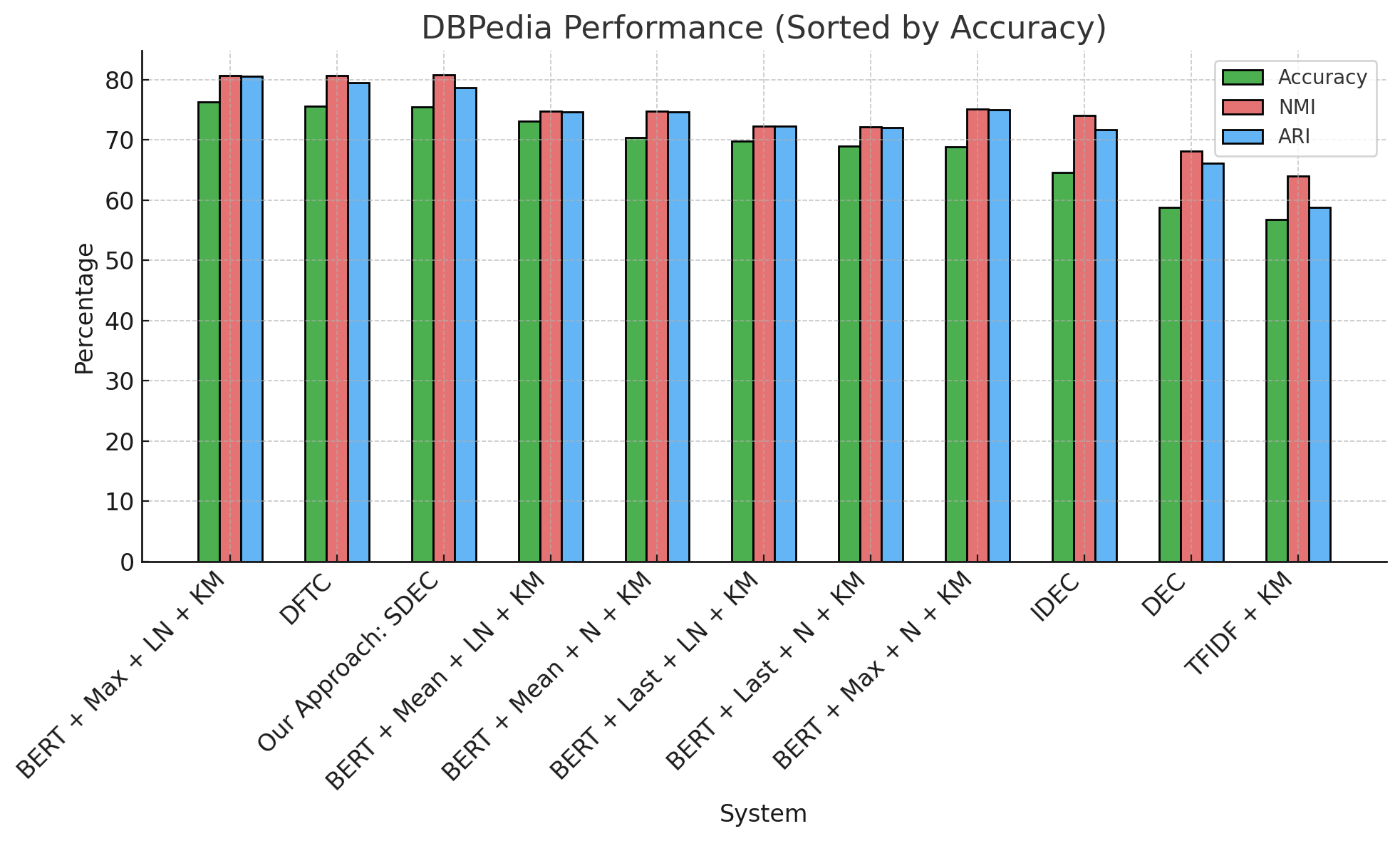}
    }
    \hfill
    \subfloat[Reuters 2 Performance (Sorted by Accuracy)\label{fig:reuters2_performance}]{%
        \includegraphics[width=0.45\textwidth]{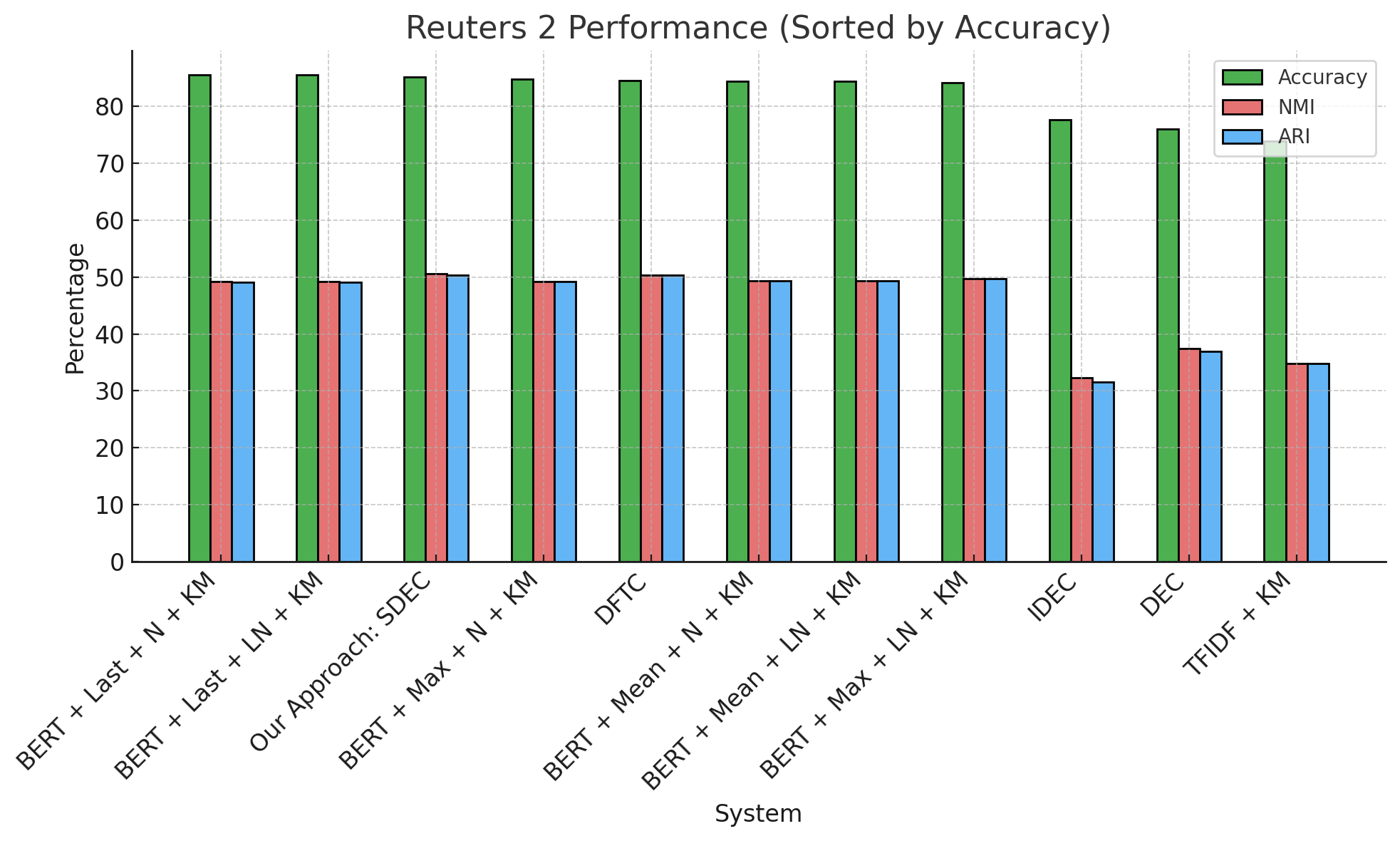}
    }
    \\
    \subfloat[Reuters 5 Performance (Sorted by Accuracy)\label{fig:reuters5_performance}]{%
        \includegraphics[width=0.45\textwidth]{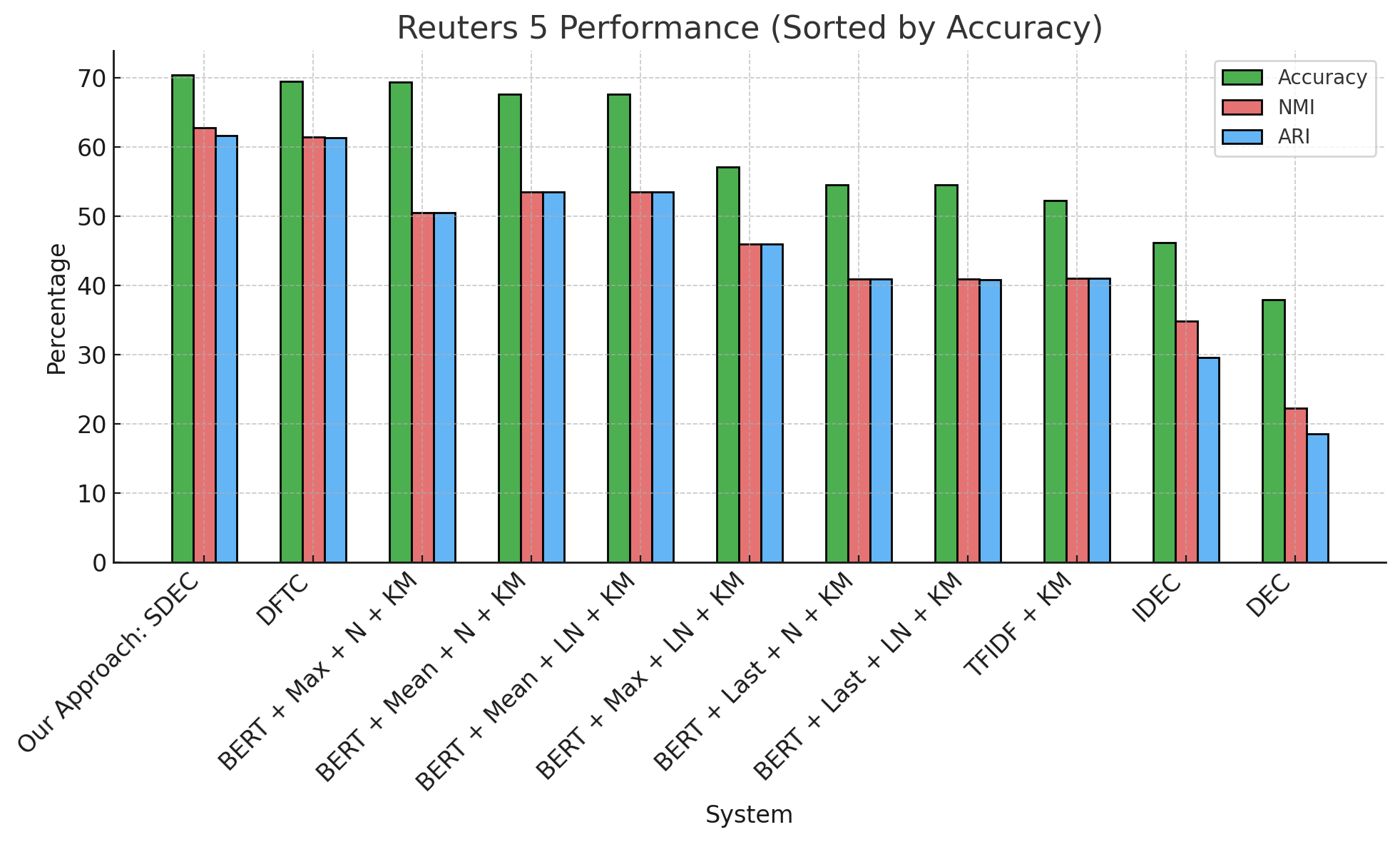}
    }
    \caption{Clustering performance comparison across different datasets.}
    \label{fig:performance_comparison}
\end{figure*}

\begin{table*}[h!]
\centering
\caption{Clustering Results Across Datasets (Best Results in Bold)}
\label{tab:result_comparison}
\renewcommand{\arraystretch}{1.2}
\setlength{\tabcolsep}{4pt}
\small
\resizebox{\textwidth}{!}{%
\begin{tabular}{|c|c@{\hspace{5pt}}c@{\hspace{5pt}}c|c@{\hspace{5pt}}c@{\hspace{5pt}}c|c@{\hspace{5pt}}c@{\hspace{5pt}}c|c@{\hspace{5pt}}c@{\hspace{5pt}}c|c@{\hspace{5pt}}c@{\hspace{5pt}}c|}
\hline
\textbf{System} & \multicolumn{3}{c|}{\textbf{AG News}} & \multicolumn{3}{c|}{\textbf{Yahoo! Answers}} & \multicolumn{3}{c|}{\textbf{DBPedia}} & \multicolumn{3}{c|}{\textbf{Reuters 2}} & \multicolumn{3}{c|}{\textbf{Reuters 5}} \\ \cline{2-16} 
 & \textbf{Accuracy} & \textbf{NMI} & \textbf{ARI} & \textbf{Accuracy} & \textbf{NMI} & \textbf{ARI} & \textbf{Accuracy} & \textbf{NMI} & \textbf{ARI} & \textbf{Accuracy} & \textbf{NMI} & \textbf{ARI} & \textbf{Accuracy} & \textbf{NMI} & \textbf{ARI} \\ \hline

NMF + TFIDF + KM                  & 0.301 & 0.022 & 0.006 & 0.305 & 0.180 & 0.081 & 0.513 & 0.584 & 0.246 & 0.734 & 0.342 & 0.214 & 0.435 & 0.303 & 0.031 \\
LSA + TFIDF + KM                  & 0.301 & 0.023 & 0.006 & 0.344 & 0.102 & 0.207 & 0.492 & 0.589 & 0.224 & 0.731 & 0.339 & 0.209 & 0.437 & 0.303 & 0.027 \\

LSA + TF-IDF + Spherical KM & 0.391 & 0.103 & 0.115 & 0.417 & 0.226 & 0.117 & 0.648 & 0.666 & 0.542 & 0.749 & 0.173 & 0.246 & 0.618 & 0.372 & 0.361 \\ 
NMF + TF-IDF + Spherical KM & 0.423 & 0.125 & 0.128 & 0.33  & 0.188 & 0.137 & 0.626 & 0.646 & 0.516 & 0.777 & 0.252 & 0.308 & 0.605 & 0.348 & 0.270 \\
TF-IDF + Spherical KM       & 0.670 & 0.482 & 0.482 & 0.436 & 0.294 & 0.215 & 0.575 & 0.699 & 0.515 & 0.764 & 0.380 & 0.277 & 0.645 & 0.516 & 0.398 \\

TFIDF + KM                      & 0.568 & 0.33  & 0.30  & 0.319 & 0.187 & 0.165 & 0.567 & 0.64  & 0.588 & 0.739 & 0.348 & 0.348 & 0.522 & 0.41  & 0.41 \\
TFIDF + EFCM                    & 0.578 & 0.297 & 0.297 & 0.252 & 0.125 & 0.1236 & - & - & - & - & - & - & - & - & - \\
TFIDF + DEC                     & 0.7211 & 0.386 & 0.413 & 0.374 & 0.2302 & 0.228 & - & - & - & - & - & - & - & - & - \\
TFIDF + IDEC                    & 0.745 & 0.425 & 0.457 & 0.397 & 0.224 & 0.147 & - & - & - & - & - & - & - & - & - \\

DEC                            & 0.8019 & 0.538 & 0.565 & 0.487 & 0.3019 & 0.247 & 0.588 & 0.681 & 0.661 & 0.76 & 0.374 & 0.369 & 0.379 & 0.222 & 0.185 \\
IDEC                           & 0.8038 & 0.538 & 0.568 & 0.475 & 0.297 & 0.2339 & 0.646 & 0.74 & 0.717 & 0.777 & 0.323 & 0.315 & 0.461 & 0.348 & 0.295 \\
BERT + Last + N + KM           & 0.354 & 0.057 & 0.056 & 0.183 & 0.05 & 0.049 & 0.69 & 0.721 & 0.72 & \textbf{0.856} & 0.492 & 0.491 & 0.545 & 0.409 & 0.409 \\
BERT + Last + LN + KM          & 0.353 & 0.056 & 0.055 & 0.193 & 0.059 & 0.057 & 0.698 & 0.723 & 0.723 & \textbf{0.856} & 0.492 & 0.491 & 0.545 & 0.409 & 0.408 \\
BERT + Max + LN + KM           & 0.827 & 0.569 & 0.569 & 0.336 & 0.178 & 0.176 & \textbf{0.763} & 0.806 & 0.805 & 0.842 & 0.497 & 0.497 & 0.571 & 0.459 & 0.459 \\
BERT + Max + N + KM            & 0.823 & 0.564 & 0.563 & 0.26  & 0.122 & 0.12  & 0.688 & 0.751 & 0.75  & 0.848 & 0.492 & 0.492 & 0.693 & 0.505 & 0.505 \\
BERT + Mean + N + KM           & 0.804 & 0.539 & 0.538 & 0.424 & 0.277 & 0.276 & 0.704 & 0.747 & 0.746 & 0.845 & 0.494 & 0.494 & 0.676 & 0.535 & 0.535 \\
BERT + Mean + LN +KM           & 0.804 & 0.538 & 0.537 & 0.356 & 0.213 & 0.212 & 0.731 & 0.747 & 0.746 & 0.845 & 0.494 & 0.494 & 0.676 & 0.535 & 0.535 \\
BERT + Max + LN + EFCM         & 0.778 & 0.497 & 0.497 & 0.402 & 0.217 & 0.162 & - & - & - & - & - & - & - & - & - \\
DFTC                           & 0.851 & \textbf{0.626} & 0.624 & 0.526 & 0.356 & \textbf{0.352} & 0.756 & 0.806 & \textbf{0.795} & 0.846 & 0.503 & 0.503 & 0.695 & 0.614 & 0.613 \\

\textbf{SDEC (0\% MSE \& 100\% CSL)}    & 0.775 & 0.44  & 0.42  & 0.475 & 0.35  & 0.248 & 0.744 & 0.772 & 0.776 & 0.842 & 0.495 & 0.468 & 0.632 & 0.52  & 0.46  \\
\textbf{SDEC (100\% MSE \& 0\% CSL)}      & 0.832 & 0.574 & 0.607 & 0.524 & \textbf{0.393} & 0.289 & 0.729 & 0.788 & 0.661 & 0.845 & 0.493 & 0.476 & 0.644 & 0.54  & 0.51  \\
\textbf{SDEC}      & \textbf{0.857} & 0.625 & \textbf{0.628} & \textbf{0.5363} & 0.363 & 0.348 & 0.755 & \textbf{0.808} & 0.787 & 0.852 & \textbf{0.506} & \textbf{0.504} & \textbf{0.704} & \textbf{0.627} & \textbf{0.616} \\

\hline
\end{tabular}%
}
\end{table*}

Using a combination of MSE and Cosine Similarity Loss with dynamic weight allocation, our suggested SDEC clustering system achieves notable improvements in the final stage (Figure \ref{fig:cluster_results} (c)). The preservation of the embeddings' magnitude and semantic similarity is guaranteed by this combined loss throughout the reconstruction process. Furthermore, by reassigning points based on their similarity to cluster centroids, the semantic refinement step enhances cluster assignments even more. This stage attains an accuracy (ACC) of 82.26, an NMI of 53.04, and an ARI of 56.5. There is very little visual overlap between different clusters, indicating a much stronger visual separation between them. Since Cluster 4 is situated on the other side of the 3D graph, it is not visible in this particular view. 

To ensure a fair comparison, the UMAP projections into 3D space for visualization were performed using the same parameters at every stage. The significant enhancement in clustering performance, as observed through numerical and visual means, demonstrates the effectiveness of our suggested methodology.

\subsection{Results from Comparative Analysis}

In this subsection, we present a comparative analysis of our SDEC clustering system with other systems across multiple datasets, including \textit{AG News, Yahoo! Answers, DBPedia, Reuters 2, and Reuters 5}. As noted in previous studies, many clustering approaches either do not specify the number of samples used or employ relatively small subsets of the datasets, which limits their applicability to large-scale, real-world clustering tasks. For example, studies such as Subakti et al. \cite{subakti2022bert} used only 4000 samples from \textit{AG News} and 10,000 samples from \textit{Yahoo! Answers}. On the other hand, our method makes use of much larger subsets of these datasets, guaranteeing a more comprehensive and reliable assessment.

Furthermore, some of the experimental results presented in this analysis were adapted from studies conducted by Subakti et al. and Guan et al. \cite{subakti2022bert, guan2020deep}. For the \textit{AG News} and \textit{Yahoo! Answers} datasets, however, we conducted our own experiments for the DEC and IDEC approaches, which demonstrated improvements over the results reported by the aforementioned authors. Therefore, we have included our results in the comparison table specifically for these two datasets. Certain approaches were not tested for the \textit{DBPedia, Reuters 2,} and \textit{Reuters 5} datasets, leading to empty entries in the corresponding sections of our comparative results table in Table \ref{tab:result_comparison}.

Moreover, other clustering techniques, like those explored by Agarap \cite{agarap2021text}, used semi-supervised methods by augmenting embeddings with labeled training data, thus disqualifying them from direct comparison with our fully unsupervised methodology. Consequently, such systems were excluded from our evaluation. However, we included other relevant studies such as Subakti et al. \cite{subakti2022bert}, which used unsupervised approaches like DEC, IDEC, and other K-Means-based clustering techniques, albeit on smaller subsets. Each experiment was run 20 times with random initializations to guarantee the accuracy and robustness of our findings, and the best clustering results were reported.

In the comparison:
\begin{itemize}
    \item \textbf{LSA} refers to Latent Semantic Analysis
    \item \textbf{NMF} refers to Non-negative Matrix Factorization
    \item \textbf{DEC} refers to Deep Embedded Clustering
    \item \textbf{IDEC} refers to Improved Deep Embedded Clustering
    \item \textbf{DFTC} refers to Deep Feature Based Text Clustering
    \item \textbf{TFIDF} refers to Term Frequency Inverse Data Frequency
    \item \textbf{N} refers to standard normalization techniques applied to embeddings.
    \item \textbf{LN} refers to Layer normalization techniques applied to embeddings.
    \item \textbf{Mean} refers to the mean pooling strategy of the embeddings.
    \item \textbf{Max} refers to the max pooling strategy of the embeddings. 
    \item \textbf{Last} refers to the last pooling strategy of the embeddings
    \item \textbf{KM} stands for K-means clustering algorithm.
    \item \textbf{EFCM} is Eigenspace-based Fuzzy C-means, a clustering method combining fuzzy logic and eigenspace transformations. \cite{muliawati2017eigenspace}
\end{itemize}

In addition to neural-based methods, we evaluated traditional topic modeling approaches paired with clustering algorithms, such as \textit{NMF + TF-IDF + K-Means} and \textit{LSA + TF-IDF + Spherical K-Means}. As shown in Table \ref{tab:result_comparison}, these classical baselines performed worse than neural network-based models. This underscores the superiority of deep embedding approaches for capturing semantic relationships in high-dimensional text data.

Our SDEC system demonstrated consistent superiority, obtained a notable 85.7\% accuracy, 62.5\% NMI, and 62.8\% ARI for the \textit{AG News} dataset. This is noticeably better than the next best-performing model, DFTC, which achieved a marginally higher NMI of 62.6\% but had slightly lower accuracy (85.1\%) and ARI (62.4\%). Several models, such as DEC and IDEC, on the other hand, trailed significantly, with accuracy levels below 81\% and NMI values approximately 53.8\%. 

Our system's resilience is demonstrated by our SDEC approach's 53.63\% accuracy, 36.3\% NMI, and 34.8\% ARI on the \textit{Yahoo! Answers} dataset, which again outperforms the rest of the field. The next-best system for this dataset, DFTC, obtained similar but still lower accuracy of 52.6\%, NMI of 35.6\%, and ARI of 35.2\%. Furthermore proving the consistency of our methodology in unsupervised clustering tasks, the models utilizing BERT with K-Means or DEC fared significantly worse, with accuracies ranging between 26\% and 48.7\%.

By achieving 75.5\% accuracy, 80.8\% NMI, and 78.7\% ARI for the \textit{DBPedia} dataset, SDEC once again demonstrated its resilience, closely matching the results of DFTC, which obtained slightly higher accuracy (75.6\%) and ARI (79.5\%) but a lower NMI (80.6\%). With more clusters and a larger range of categories, this dataset exhibits a notable increase in complexity. Our strategy handled the increased complexity well, ensuring competitive results across all metrics, by means of optimized pooling, semantic loss, and layer normalization strategies. 

The \textit{Reuters 2} dataset, due to its simplicity, allowed multiple systems to perform well. Among the best performers, our SDEC had accuracy of 85.2\%, NMI of 50.6\%, and ARI of 50.4\%. Nonetheless, the performance disparities between different systems were negligible because of the dataset's relative simplicity. Models such as BERT + Last + N + KM demonstrate this, achieving 85.6\% accuracy and 49.1\% ARI, proving that even more basic models can function well when the clustering problem is less complicated. 

Lastly, on the more challenging \textit{Reuters 5} dataset, our SDEC approach demonstrated superior performance with an accuracy of 70.4\%, 62.7\% NMI, and 61.6\% ARI, significantly outperforming other systems such as DEC and IDEC, which struggled with accuracies below 38\%. The better performance of SDEC on \textit{Reuters 5} indicates that our framework also excels in dealing with moderately complex clustering tasks, where other unsupervised methods falter.

While SDEC and DFTC may achieve similar numerical results, SDEC offers important methodological benefits. Unlike DFTC’s usage of static embeddings, SDEC fine-tunes representations during clustering using a dual-loss approach that jointly optimizes semantic alignment and reconstruction. This dynamic optimization and refinement make SDEC more robust and adaptable, especially for tasks needing deeper semantic understanding.

To visually highlight the comparative performance among advanced neural-based clustering methods, we present Figure~\ref{fig:performance_comparison}, which shows the clustering accuracy results specifically for deep learning-based techniques across all datasets. These figures (\ref{fig:ag_performance}–\ref{fig:reuters5_performance}) illustrate that the proposed SDEC framework consistently achieves strong performance compared to other deep clustering approaches, such as DEC, IDEC, and transformer-based K-means variants. Note that the figures exclude classical topic modeling-based methods and ablation study variants, which are comprehensively detailed numerically in Table~\ref{tab:result_comparison}.

To further validate our proposed loss combination (MSE + CSL) and dynamic weight allocation, we performed targeted ablation experiments summarized in Table~\ref{tab:result_comparison}. Specifically, we compared two variations of our framework: SDEC focusing exclusively on semantic alignment, where the loss emphasized angular similarity between embeddings (CSL only, 0\% MSE and 100\% CSL) and SDEC focusing exclusively on embedding regeneration, where the loss prioritized reconstructing the original embedding values (MSE only, 100\% MSE and 0\% CSL). The results from these experiments show that focusing solely on semantic alignment generally yields suboptimal performance across all datasets, as evidenced by decreased accuracy and ARI scores. In contrast, the embedding regeneration variant consistently outperformed semantic alignment alone, suggesting that emphasizing the faithful reconstruction of embedding values helps preserve more discriminative features for clustering. However, the full SDEC model, which dynamically balances both objectives, clearly surpasses both single-loss configurations, particularly on more complex datasets. These findings underscore the effectiveness and necessity of the integrated loss function design in the SDEC framework.

Table~\ref{tab:result_comparison} provides a comprehensive summary of clustering performance results across all evaluated methods and datasets, including both classical topic modeling baselines, ablation variants of our proposed method, and state-of-the-art deep clustering frameworks. This extensive comparison clearly demonstrates that our proposed SDEC system consistently outperforms other unsupervised methods across all metrics. Whether the dataset complexity is high, moderate, or low, SDEC maintains superior accuracy, NMI, and ARI scores, highlighting its effectiveness and robustness across diverse clustering scenarios.



\section{Conclusions and Discussions}

In order to improve text clustering accuracy, this research presented Semantic Deep Embedded Clustering (SDEC), a novel framework for unsupervised text clustering that combines semantic refinement with a combined semantic loss. With the use of transformer-based BERT embeddings and an optimized autoencoder, SDEC has proven to perform better on a range of datasets with varying degrees of complexity.
\titlespacing*{\subsection}{0pt}{1.25ex plus 1ex minus .2ex}{0.75ex plus .2ex}
\subsection{Summary of Findings}
Our results show that SDEC performs consistently better than conventional clustering systems. Among the notable accomplishments are accuracy rates of 85.7\% on \textit{AG News} and 53.63\% on the more complex \textit{Yahoo! Answers}, which are higher than conventional approaches on these datasets. Additionally, SDEC demonstrated robust performance on \textit{DBPedia}, \textit{Reuters 2}, and \textit{Reuters 5}, demonstrating the model's effectiveness in managing a variety of text corpora and obtaining good metrics on all datasets.
\titlespacing*{\subsection}{0pt}{1.25ex plus 1ex minus .2ex}{0.75ex plus .2ex}
\subsection{Key Insights}
The SDEC framework's effectiveness stems from:
\begin{itemize}
    \item The integration of semantic and distributional losses, ensuring deep semantic preservation within the clustering process.
    \item Semantic refinement techniques that enhance clustering definitions, particularly beneficial in complex datasets like \textit{DBPedia} and \textit{Yahoo! Answers}.
    \item Consistent performance across various datasets, illustrating the method's adaptability and robustness.
\end{itemize}

\subsection{Contributions}
SDEC's primary contributions include:
\begin{itemize}
    \item An enhanced clustering framework that integrates semantic loss and semantic refinement for improved clustering accuracy and semantic integrity.
    \item Demonstrated improvements in clustering performance across five major datasets, showcasing versatility.
    \item Establishment of a robust methodology that surpasses traditional DEC and IDEC approaches, especially in complex clustering scenarios.
\end{itemize}

\subsection{Limitations}
Clustering is fundamentally an NP-hard problem \cite{dasgupta2008hardness}, implying that no polynomial-time algorithm can guarantee a globally optimal solution. Therefore, the concept of a definitive, universally “correct” clustering outcome is inherently ambiguous in unsupervised learning, requiring reliance on heuristic and approximation-based methods. Although SDEC demonstrates strong performance, certain limitations remain. Firstly, the model assumes prior knowledge of the number of clusters, which might not always be realistic in open-domain or exploratory scenarios. Secondly, transformer-based embeddings, despite their proven effectiveness, incur substantial computational costs, potentially limiting the scalability of SDEC to extremely large datasets without specialized computational infrastructure or optimization. Thirdly, SDEC inherently lacks interpretability which is a common limitation of complex deep learning-based clustering models, making it challenging to directly explain why certain documents are grouped together, which may be a drawback in domains requiring transparency and explainability. Furthermore, the framework introduces numerous hyperparameters (e.g., loss weights, refinement thresholds, learning rates), and optimal tuning may require significant computational resources and domain expertise, potentially creating a barrier for new users. We also note that, the current experimental validation has been conducted exclusively on English-language datasets using BERT embeddings. While this setup is effective for the present benchmarks, the applicability of SDEC to multilingual contexts or with alternative embedding models remains unexplored and is an important area for future research. Lastly, the SDEC framework depends on the initialization of cluster centers using the k-means++ strategy. While superior to random initialization, this approach remains stochastic and can influence final clustering outcomes. As a consequence, the clustering quality might vary across different runs, even with fixed hyperparameters, due to sensitivity in the initial cluster assignments within the latent space.

\subsection{Future Directions}
Potential areas for future research include:
\begin{itemize}
    \item Exploring scalability to larger and more heterogeneous datasets to further test SDEC's effectiveness. To address scalability limitations arising from computationally expensive transformer-based embeddings, methods such as embedding compression, efficient retrieval strategies, or lightweight transformer models should be explored.
    \item Investigating hybrid models that merge unsupervised and supervised learning to refine feature representations and clustering outcomes. In particular, such hybrid approaches offer a promising path for addressing cluster imbalance \cite{lin2017clustering,truica2017classification}, as supervised guidance and imbalance-aware loss functions can help ensure that minority classes are better represented and more accurately clustered.
    \item Developing robust initialization methods or ensemble clustering approaches to mitigate sensitivity issues related to stochastic initialization of cluster centers. Advanced initialization strategies, meta-learning, or consensus clustering methods could provide more stable and consistent clustering performance.
    \item Extending SDEC applications to real-world tasks such as document categorization and recommendation systems, to validate its practical utility. Evaluating performance in realistic scenarios will inform improvements in computational efficiency, interpretability, and practical robustness.
    \item Addressing the challenge of multimodal data, where text is combined with other modalities such as images, audio, or structured metadata. Adapting SDEC to jointly process and align heterogeneous feature spaces will require extending the framework to incorporate modality-specific encoders and cross-modal fusion mechanisms, enabling more comprehensive clustering of complex, real-world datasets.
\end{itemize}

Overall, SDEC sets a new standard for unsupervised text clustering, effectively addressing both the challenge of semantic preservation and the need for high clustering accuracy. Future enhancements and applications of this framework have the potential to significantly advance the field of text clustering.

\bibliographystyle{IEEEtran}
\bibliography{references}
\section*{Biography}
\vspace{-25.5mm}
\begin{IEEEbiography}[{\includegraphics[width=1.05in,height=1.3125in,clip,keepaspectratio]{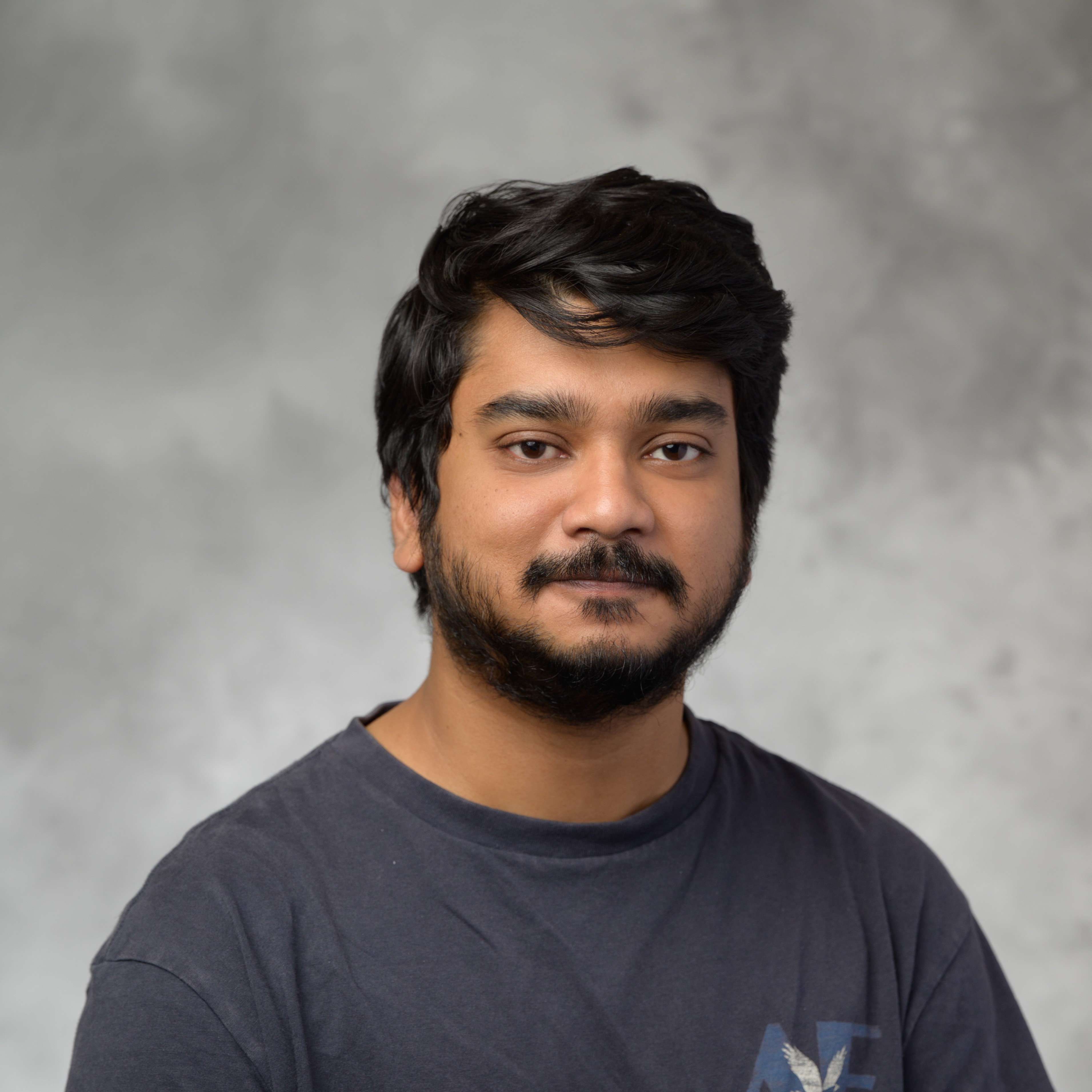}}]{Mohammad Wali Ur Rahman}
graduated with a Master of Science degree in Electrical and Computer Engineering from the University of Arizona in 2023 and is currently pursuing his Ph.D. in the same field at the same institution. He has been engaged as a Research Assistant at the Autonomic Computing Lab, a branch of the NSF-CAC (NSF Center for Autonomic Computing). His academic and research pursuits are centered around Text Mining \& Analysis, Natural Language Processing, Machine Learning, Neural Networks, Artificial Intelligence and cybersecurity.
\end{IEEEbiography}
\vspace{-25mm} 

\begin{IEEEbiography}[{\includegraphics[width=1.05in,height=1.3125in,clip,keepaspectratio]{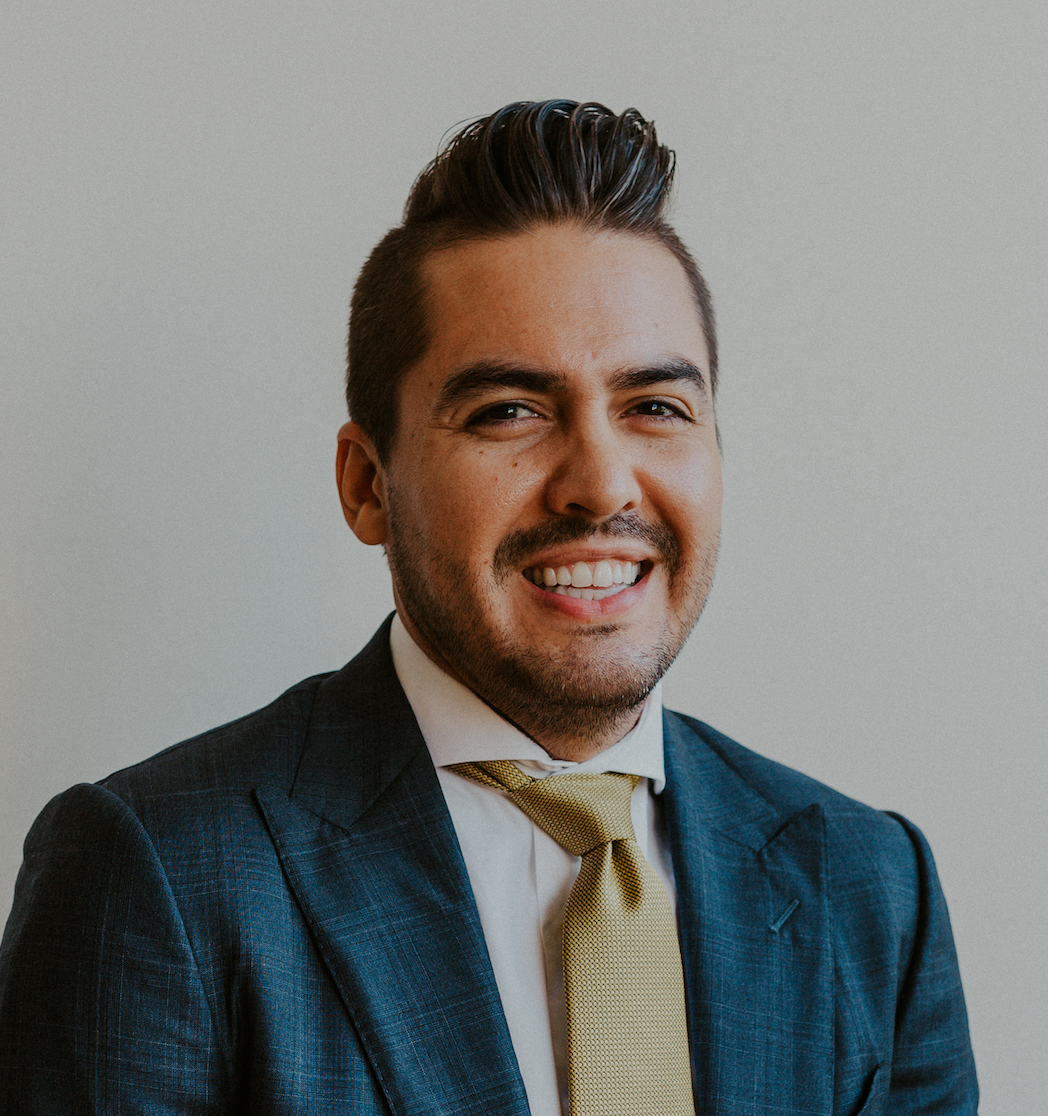}}]{Ric Nevarez}
is the Chief Technology Officer at Trust Web. He graduated with a Bachelor of Science degree in Nutritional Biochemistry from Brigham Young University in 2014. With over 14 years of experience in the artificial intelligence and natural language processing industry, Ric Nevarez has been a strong advocate for sustainable and ethical AI practices. As an experienced product manager and entrepreneur, he has co-founded several successful AI startups, securing substantial funding and developing innovative AI products. 
\end{IEEEbiography}
\vspace{-15mm} 
\newpage
\begin{IEEEbiography}[{\includegraphics[width=1.05in,height=1.3125in,clip,keepaspectratio]{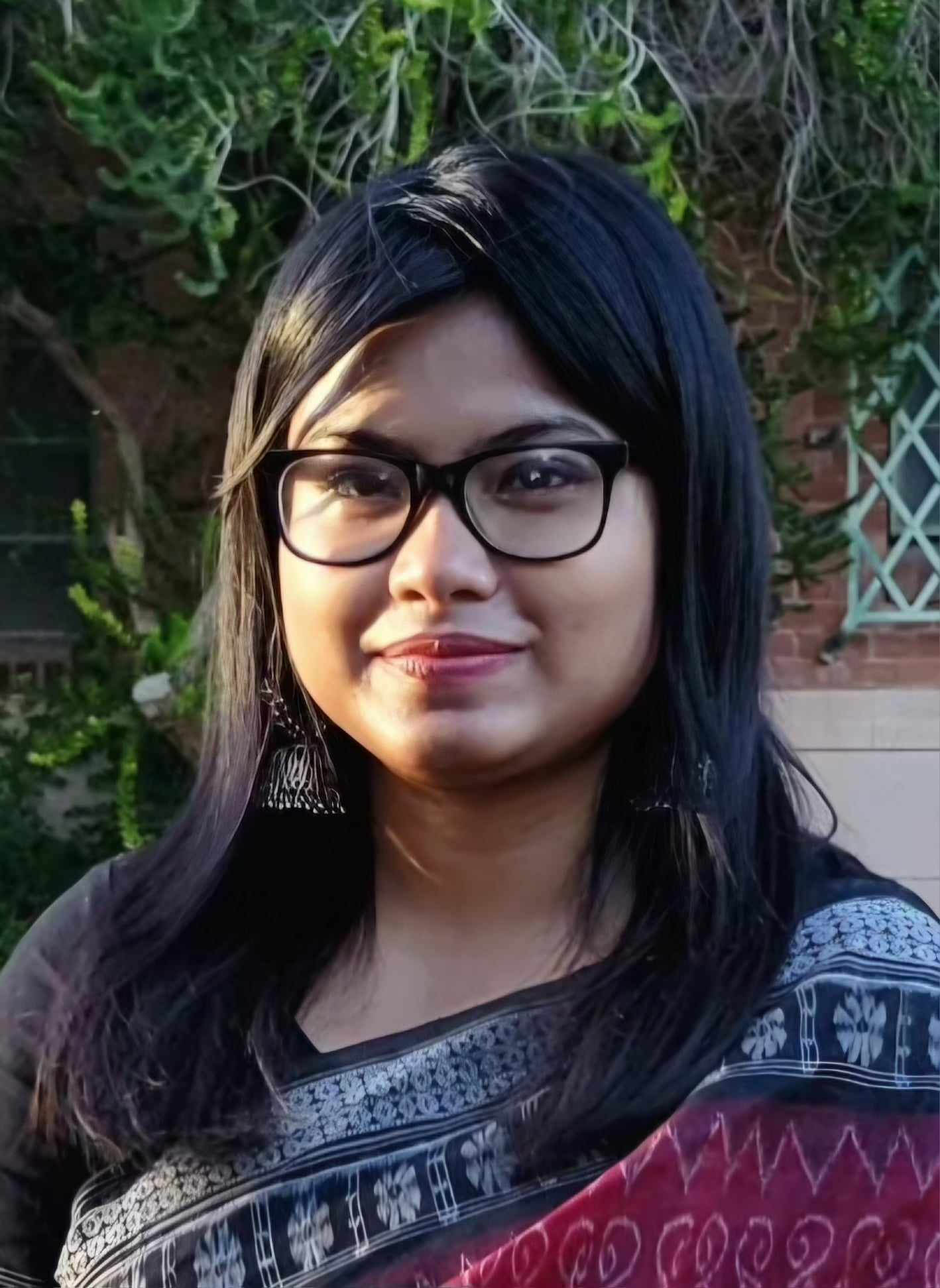}}]{Lamia Tasnim Mim}
 is a graduate assistant at New Mexico State University, where she earned her Master's in Computer Science (2023). She specializes in Artificial Intelligence, Machine Learning, and Natural Language Processing, with a focus on developing robust and scalable deep learning models. Lamia served as a Machine Learning Engineer at Avirtek, Inc., where she developed streaming data pipelines, deployed ML models, and enhanced data recovery accuracy through advanced techniques. 
\end{IEEEbiography}
\vspace{-205mm}

\begin{IEEEbiography}[{\includegraphics[width=1.05in,height=1.3125in,clip,keepaspectratio]{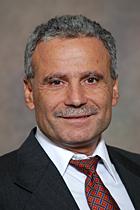}}]{Dr Salim Hariri}
(Senior Member, IEEE) received an
M.Sc. degree from Ohio State University in 1982, and a Ph.D. degree in Computer Engineering from the University of Southern California in 1986. He is a Professor in the Department of Electrical and Computer Engineering, the University of Arizona, and the Director of the NSF Center for Cloud and Autonomic Computing (NSF-CAC). His research focuses on autonomic computing, artificial intelligence, machine learning, cybersecurity, cyber resilience, and cloud security.
\end{IEEEbiography}
\vspace{-35mm} 

\clearpage

\appendices

\section{Preprocessing and Lemmatization}
\label{appendix:preprocessing}
The preprocessing step is critical in ensuring that the textual data is clean, consistent, and suitable for further analysis. This section describes the preprocessing steps done on the text data: lemmatizing the tokens, cleaning the text with regular expressions, removing URLs, and expanding contractions. The aim is to convert the unprocessed text into a format that is standardized and suitable for producing embeddings.

To start, for greater consistency and clarity, the text data frequently needs to have contractions expanded. We enlarge them in the text by using a predefined dictionary of contractions.

Next, using a regular expression pattern, we preserve only alphabets and choose punctuation characters (like `!', `?', `,', and `.'). The text is made cleaner by removing all other characters.

To highlight the important passages, certain redundant words are also eliminated from the text. Following this, the text has been cleaned and tokenized into individual words, dividing the text into manageable chunks for subsequent processing.

After tokenization, more filtering is used to eliminate particular undesirable tokens that don't improve the text analysis. After that, the tokens are lemmatized to return to their base or root form, which aids in normalizing the text and simplifying the vocabulary.

After lemmatization, the normalized text is represented by joining the tokens back into a single string. These preprocessing procedures help us make sure the text data is clean, consistent, and suitable for producing high-quality embeddings. Our clustering framework's later steps are built upon this processed text data.

\hfill \break
\section{BERT Embedding Generation}
\label{appendix:bert}
We use BERT (Bidirectional Encoder Representations from Transformers) embeddings, namely the pre-trained "bert-large-cased" model, to take full advantage of the power of transformer models for text clustering. For the text data to be successfully converted into high-quality embeddings that accurately capture the text's semantic meaning, a number of steps must be taken during the generation of BERT embeddings. 

The key steps for generating BERT embeddings are as follows:

\begin{itemize}
    \item \textbf{Load BERT Tokenizer and Model}:
    \begin{itemize}
        \item Utilize the \texttt{transformers} library to load the BERT tokenizer and model.
        \item The tokenizer converts text into tokens that the BERT model can process.
    \end{itemize}

    \item \textbf{Encode Text Data}:
    \begin{itemize}
        \item For each text sample, the tokenizer encodes the text, adding special tokens and ensuring proper padding and truncation to maintain a consistent input length.
        \item The encoded inputs are passed through the BERT model to obtain token embeddings from the last hidden state.
    \end{itemize} 
    
    \item \textbf{Pooling Strategies}:
    \begin{itemize}
        \item We experimented with three different pooling strategies to derive fixed-size representations for each text sample:
        \begin{itemize}
            \item \textbf{CLS Pooling}: Taking the [CLS] token embedding, designed to capture the aggregate representation of the input text.
            \item \textbf{Mean Pooling}: Calculating the average of the token embeddings to obtain a representation that reflects the overall content of the text. This can be represented as:
            \begin{equation}
            h[k] = \frac{1}{n} \sum_{i=1}^{n} h_{ik},
            \end{equation}
            where \( h[k] \) is the \( k \)-th entry of the feature-extracted representation vector \( h \), and \( h_{ik} \) is the \( k \)-th entry of the \( i \)-th token embedding.
            \item \textbf{Max Pooling}: Taking the maximum value across all token embeddings, which helps to capture the most salient features of the text. This can be represented as:
            \begin{equation}
            h[k] = \max_{1 \leq i \leq n} h_{ik},
            \end{equation}
            where \( h[k] \) is the \( k \)-th entry of the feature-extracted representation vector \( h \).
        \end{itemize}
        \item Among these strategies, mean pooling typically yields the best results for clustering, providing a balanced representation of the text content.
    \end{itemize}

    \item \textbf{Convert to Numpy Arrays}:
    \begin{itemize}
        \item Convert the mean-pooled BERT embeddings into numpy arrays for further processing.
    \end{itemize}

    \item \textbf{Normalization}:
    \begin{itemize}
        \item We experimented with two types of normalization techniques:
        \begin{itemize}
            \item \textbf{Standard Normalization}: Applying a \texttt{StandardScaler} to ensure the embeddings are on a comparable scale. This transforms the vector representation into a vector with a norm of 1. This can be represented as:
            \begin{equation}
            \bar{h_i} = \frac{h_i}{\|h_i\|},
            \end{equation}
            where \( \bar{h_i} \) is the normalized vector.
            \item \textbf{Layer Normalization}: Ensuring the stability and consistency of the embeddings across different layers by avoiding covariate shift problems. This can be represented as:
            \begin{equation}
            \bar{h_i} = \frac{h_i - \phi_i}{\sigma_i},
            \end{equation}
            where \( \phi_i \) and \( \sigma_i \) are the mean and standard deviation of the feature vector \( h_i \), respectively.
        \end{itemize}
        \item Fit and transform the training embeddings, and transform the testing embeddings using the same scaler.
    \end{itemize}
\end{itemize}

By following these steps, we generate BERT embeddings that preserve the semantic meaning of the text and are standardized for subsequent use in the autoencoder and clustering framework. These embeddings form the foundation for the next stages of our methodology, ensuring high-quality input data for effective clustering.

\section{Hyperparameters and Experiment Settings}
\label{appendix:hyperparams}

This appendix provides a comprehensive overview of the hyperparameters and experimental configurations employed for training and evaluating the proposed Semantic Deep Embedded Clustering (SDEC) model. To ensure clarity and ease of comparison with prior models, a summary of key hyperparameter settings from the DEC and IDEC models is also included (Table \ref{tab:hyperparam_comparison}). Additionally, sensitivity analyses conducted specifically for the critical hyperparameters unique to the SDEC framework are discussed.

\subsection{Comparative Hyperparameter Overview}

Table \ref{tab:hyperparam_comparison} summarizes the principal hyperparameter choices adopted in DEC, IDEC, and our proposed SDEC framework. In alignment with established practice in deep clustering literature, nearly all hyperparameters in SDEC are selected from commonly used values in deep neural network training and transformer-based (e.g., BERT) embedding workflows. Consistent with DEC and IDEC, we deliberately avoid dataset-specific hyperparameter tuning to promote generalizability and fair comparison. Nevertheless, SDEC introduces several additional design elements, such as enhanced regularization and loss weighting strategies, which are reflected in the table and further explored through targeted sensitivity analyses.

\begin{table*}[t]
\centering
\caption{Comparison of Key Hyperparameters Across DEC, IDEC, and SDEC}
\label{tab:hyperparam_comparison}
\resizebox{\textwidth}{!}{%
\begin{tabular}{|l|c|c|c|}
\hline
\textbf{Category} & \textbf{DEC} & \textbf{IDEC} & \textbf{SDEC (Ours)} \\
\hline
Architecture & d–500–500–2000–10 & Same + mirrored decoder & d–2048–1024–512–256–128, symmetric decoder \\
Activation Function & Not stated & ReLU & SeLU, linear at output \\
Regularization & None specified & None specified & L2 regularization \\
Embedding Dimension & 10 & 10 & 128 \\
Batch Size & 256 & 256 & 16 (AE), 32 (Clustering) \\
Autoencoder Optimizer & SGD & Adam/SGD & Adam \\
AE Learning Rate & 0.1 & Varied: 0.1 to $1\times10^{-4}$ & Varied: $1\times10^{-3}$ to $1\times10^{-5}$ \\
Epochs & 100,000 iterations & Not specified & 100–1500 epochs (AE), 20,000 iterations (Clustering) \\
K-Means Restarts & 20 & 20 & 20 - 2000 (KMeans++) \\
Clustering Loss & KL divergence & KL + Reconstruction & KL + Semantic Reconstruction \\
Clustering Optimizer & SGD (0.1) & Adam/SGD & SGD (0.01, momentum 0.9) \\
Clustering Batch Size & 256 & 256 & 32 \\
KL α (Student’s t-kernel) & Implicit & Not stated & Explicit α = 1.0 \\
KL Convergence Threshold & δ = 0.001 & δ = 0.001 & δ = 0.001 \\
Update Interval & Not stated & Not stated & varied between 2, 5, 10, 50 \& 100 iterations \\
Cluster Coefficient & N/A & 0.1 & 0.1 \\
Cluster Refinement Threshold & N/A & N/A & λ threshold $\in$ [0.1, 0.7] \\
\hline
\end{tabular}}
\end{table*}

\subsection{Autoencoder Hyperparameters}

The autoencoder architecture plays a pivotal role in our experiments. It employs L2 regularization and SeLU activation functions throughout its hidden layers. The encoder comprises four fully connected layers with 2048, 1024, 512, and 256 units, respectively, followed by a bottleneck (latent) layer of 128 units to capture the key features of the input embeddings. The decoder mirrors the encoder’s architecture with layers of 256, 512, 1024, and 2048 units, and the output layer uses a linear activation function for reconstruction. Training of the autoencoder is performed using the Adam optimizer, with a learning rate varied between $1\times10^{-3}$ and $1\times10^{-5}$, a batch size of 16, and a composite reconstruction loss combining Mean Squared Error (MSE) and Cosine Similarity Loss (CSL). The number of training epochs ranges from 100 to 1500, depending on the dataset size and complexity.

\subsection{Clustering and Fine-Tuning Hyperparameters}

Cluster centroids are initialized using the K-Means++ algorithm, with initialization runs ranging from $2\times10^{1}$ to $2\times10^{3}$. Given that clustering outcomes from K-Means-based methods are highly sensitive to centroid initialization, employing a higher number of initializations typically yields more robust and stable cluster configurations, providing an improved foundation for subsequent optimization steps. For clustering optimization, we employ stochastic gradient descent (SGD) with a learning rate of 0.01 and momentum of 0.9, a widely adopted setting that provides effective convergence and avoids local minima. Following the recommendations established by the authors of IDEC, the clustering coefficient was set to 0.1 to suitably balance the KL divergence and semantic reconstruction objectives, facilitating the generation of semantically coherent clusters. The clustering kernel employs a Student’s $t$-distribution with an explicit parameter $\alpha$ set to 1.0, enabling the clustering process to effectively manage varying cluster densities and outliers in the embedding space. The convergence criteria were defined through a KL divergence threshold ($\delta$) of 0.001 and a maximum delta-label tolerance of 0.001, ensuring that the algorithm halts only upon achieving substantial stability in cluster assignments. During the clustering phase, cluster assignments and target distributions were updated at intervals experimentally varied among 2, 5, 10, 50, and 100 iterations, with the maximum number of clustering iterations capped at 20,000 using a batch size of 32.

\begin{figure*}[t]
    \centering
    \subfloat[AE-phase loss-weight variations (MSE vs CSL)\label{fig:ae_loss_sweep}]{
        \includegraphics[width=0.485\linewidth]{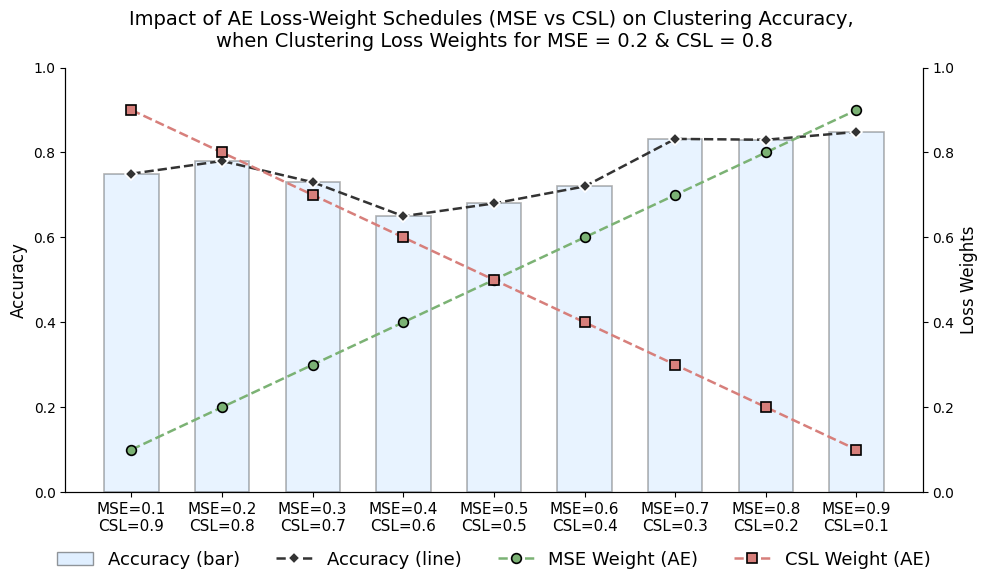}
    }
    \hfill
    \subfloat[Clustering-phase loss-weight variations (MSE vs CSL)\label{fig:cluster_loss_sweep}]{
        \includegraphics[width=0.485\linewidth]{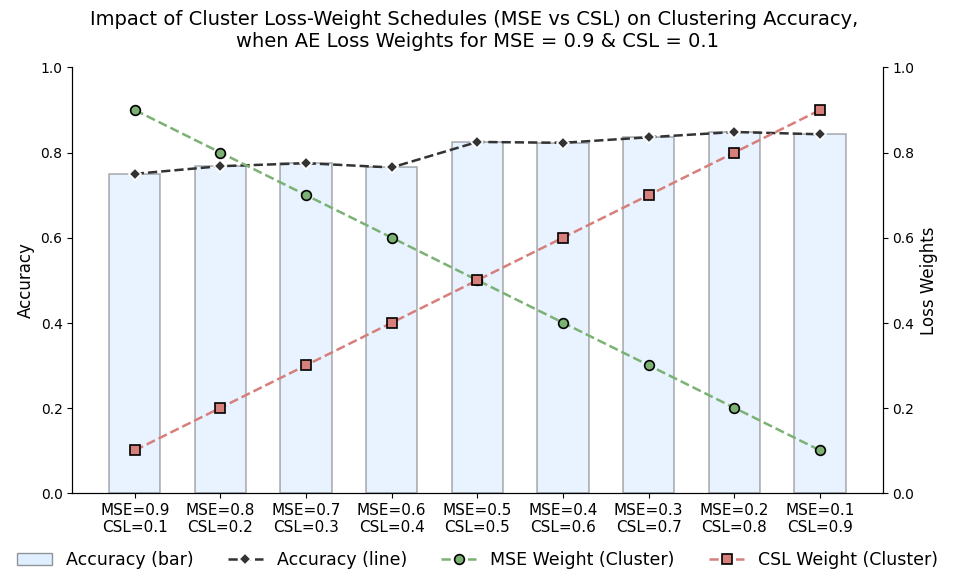}
    }
    \caption{Impact of loss-weight scheduling on clustering accuracy: (a) Autoencoder (AE) phase and (b) clustering phase.}
    \label{fig:loss_weight_sensitivity}
\end{figure*}

\subsection{Loss Weight Sensitivity Experiments}

Although the SDEC algorithm employs dynamically allocated reconstruction loss weights, controlled experiments were conducted to explicitly examine how fixed variations in the Mean Squared Error (MSE) and Cosine Similarity Loss (CSL) weights influence clustering performance. These experiments involved independent runs on the AG News dataset, without targeting optimal performance explicitly, but rather aiming to investigate general sensitivity trends in clustering accuracy.

\begin{itemize}
    \item \textbf{Autoencoder Phase (AE) Loss Weight Sweep:} During the autoencoder pretraining phase, we systematically varied the loss weight ratio between MSE and CSL while maintaining fixed weights (MSE=0.2, CSL=0.8) during the clustering phase. Results (Figure \ref{fig:ae_loss_sweep}) revealed that clustering accuracy notably declines when MSE and CSL weights are balanced or nearly balanced (e.g., 0.5–0.5 or 0.6–0.4). A plausible explanation for this accuracy dip is that balanced weighting may cause conflicting objectives between accurate embedding reconstruction (MSE) and semantic alignment (CSL), reducing the clarity and discriminative capability of the learned embeddings. Conversely, when one loss dominates, particularly the MSE, the autoencoder tends to learn more stable and consistent latent representations, leading to higher accuracy.
    
    \item \textbf{Clustering Phase Loss Weight Sweep:} For this experiment, the autoencoder training weights were held constant (MSE=0.9, CSL=0.1), emphasizing embedding stability. Clustering phase loss weights, however, were varied. Results (Figure \ref{fig:cluster_loss_sweep}) clearly indicated improved clustering accuracy as the CSL weight increased. This outcome highlights the importance of semantic alignment in the clustering stage, where a stronger CSL emphasis encourages cluster centroids to align more meaningfully with semantically similar groups of embeddings, thereby improving overall cluster quality.
\end{itemize}

The rationale behind setting constant weights of MSE=0.2 and CSL=0.8 in the clustering phase (during the AE-phase weight variations) was to prioritize semantic similarity during clustering, allowing us to assess whether stable embeddings from the autoencoder training phase influenced clustering effectiveness. Conversely, we chose fixed autoencoder-phase weights (MSE=0.9, CSL=0.1) during the clustering-phase variations to maintain consistent and stable initial embeddings, enabling a clear analysis of how varying clustering weights impact clustering performance.

These analyses underscore the robustness of the dynamic weighting strategy originally adopted by the SDEC framework, demonstrating clear insights into the individual and combined contributions of reconstruction losses during autoencoder training and clustering.

\begin{figure*}[t]
    \centering
    \subfloat[\label{fig:ag_lambda_plot}]{
        \includegraphics[width=0.48\linewidth]{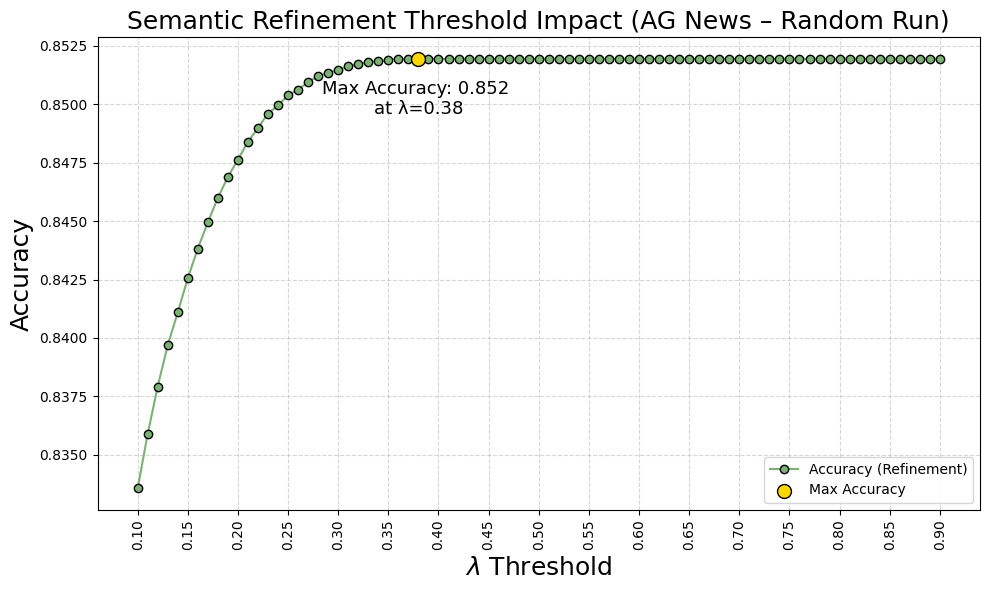}
    }
    \hfill
    \subfloat[\label{fig:r5_lambda_plot}]{
        \includegraphics[width=0.48\linewidth]{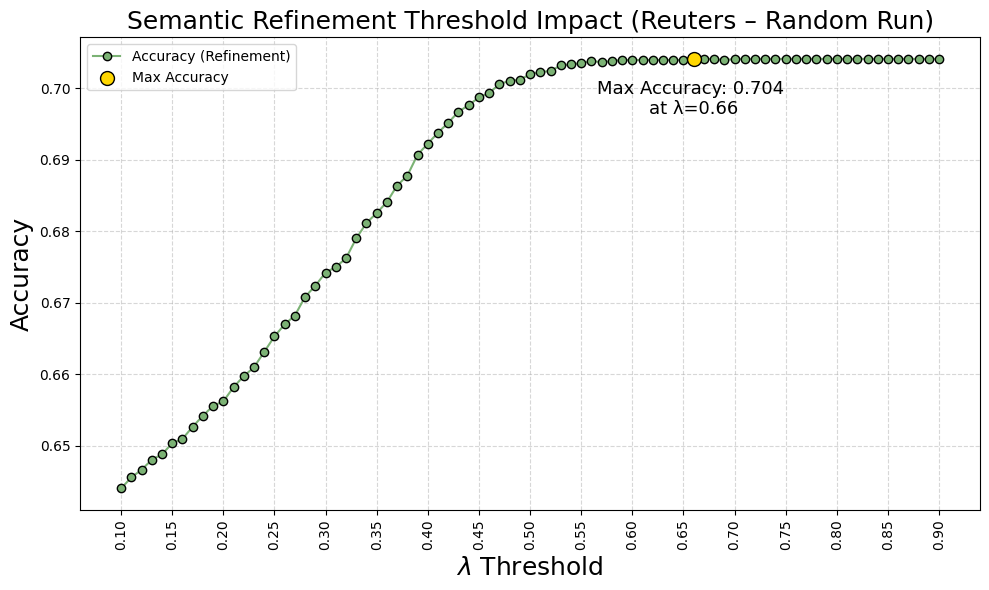}
    }
    \\
    \subfloat[\label{fig:yh_lambda_plot}]{
        \includegraphics[width=0.48\linewidth]{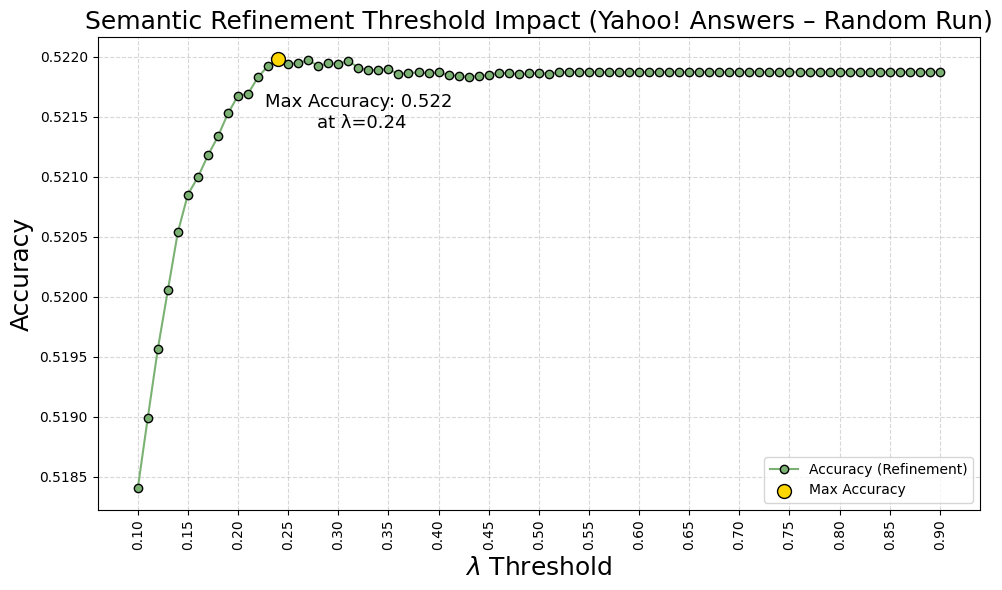}
    }
    \hfill
    \subfloat[\label{fig:db_lambda_plot}]{
        \includegraphics[width=0.48\linewidth]{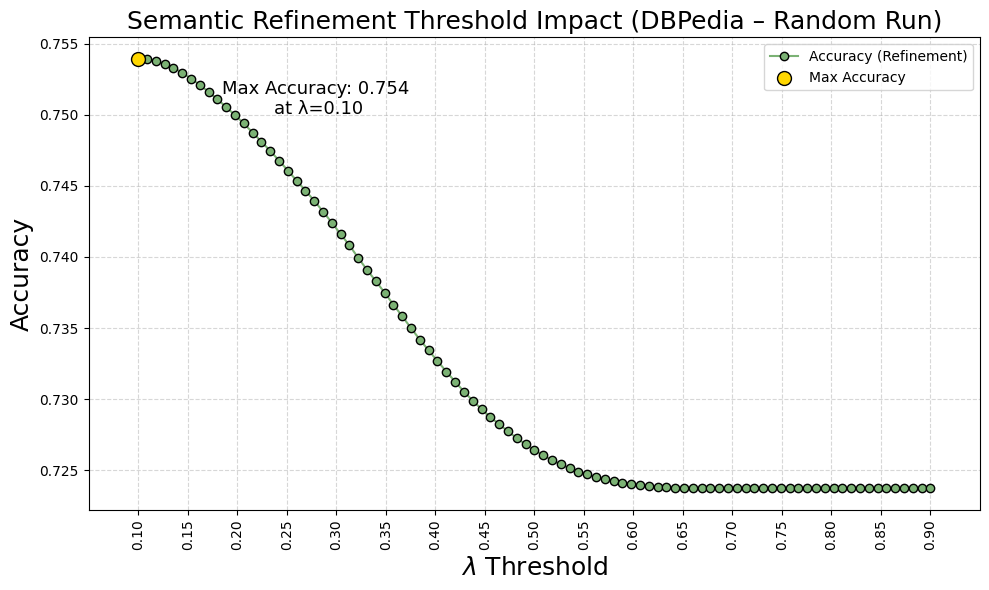}
    }
    \caption{Sensitivity analysis of the semantic refinement threshold $\lambda$ on (a) AG News, (b) Reuters 5, (c) Yahoo! Answers, and (d) DBPedia datasets.}
    \label{fig:lambda_sensitivity_analysis}
\end{figure*}

\subsection{Cluster Refinement Hyperparameters}

The semantic refinement step introduces a refinement threshold ($\lambda$), which specifies the minimum required increase in semantic similarity for data points to be reassigned to new clusters. The main goal of this sensitivity analysis was to thoroughly understand how varying the $\lambda$ threshold influences the clustering performance, rather than achieving the optimal clustering outcome itself. To accomplish this, we conducted systematic experiments using randomly initialized models across four benchmark datasets: \textit{AG News}, \textit{Yahoo! Answers}, \textit{Reuters 5}, and \textit{DBPedia}. These experiments covered a comprehensive range of $\lambda$ values from 0.1 to 0.9, evaluated in increments of 0.01, as illustrated in Figure \ref{fig:lambda_sensitivity_analysis}.

Our results consistently demonstrate that moderate $\lambda$ values (approximately between 0.1 and 0.7) yield the best clustering outcomes. The rationale behind this finding relates directly to the nature of high-dimensional semantic embeddings, such as those generated by transformer-based models like BERT. Although initial clustering methods (e.g., K-means++) can effectively initialize clusters, they may not fully capture subtle semantic distinctions due to inherent limitations of centroid-based methods in high-dimensional feature spaces. The semantic refinement step addresses this issue by making minor adjustments to cluster assignments based on nuanced semantic similarity, thereby improving the semantic coherence within clusters.

If the refinement threshold $\lambda$ is set too low (below 0.1), the model tends to make aggressive, often unnecessary reassignments, disrupting cluster stability and coherence. Conversely, excessively high values of $\lambda$ (above 0.7) severely restrict the model’s ability to make meaningful refinements, as the threshold for reassignment becomes overly stringent. Such conservatism reduces the potential improvements in cluster quality, especially when the initial clustering quality leaves room for semantic enhancement.

Hence, based on these empirical observations, we recommend selecting a refinement threshold ($\lambda$) within the range of 0.1 to 0.7 as a sensible starting point. Since the ideal setting of this hyperparameter depends on the specific semantic and structural characteristics of the dataset at hand, researchers and practitioners should consider $\lambda$ a tunable hyperparameter. The refinement threshold must therefore be selected with careful consideration of the dataset’s complexity, semantic characteristics, and the performance expectations of the clustering system.

Ultimately, the semantic refinement process significantly enhances the quality of clusters, provided the threshold $\lambda$ is carefully tuned to balance sensitivity and specificity in cluster reassignments. This controlled adjustment ensures robust and meaningful semantic alignment, thereby improving overall cluster coherence and reliability across diverse textual datasets.

\end{document}